\documentclass[9pt]{article}
\usepackage[a4paper,left=2.5cm, right=2.5cm, top=3cm, bottom=3cm]{geometry} % Adjust margins
\usepackage[english]{babel}
\usepackage{graphicx} % Required for inserting images
\usepackage{amsmath,amsfonts}
\usepackage{mathtools}
\usepackage{cite}
\usepackage{hyperref}% add hypertext capabilities
\usepackage[ruled,vlined]{algorithm2e}

\SetCommentSty{mycommfont}
\usepackage[noabbrev,nameinlink]{cleveref}%Cref
\usepackage{orcidlink} %orcid

\usepackage[dvipsnames]{xcolor}
\usepackage{authblk}

\definecolor{mycolor1}{RGB}{24, 92, 68}

\hypersetup{
    colorlinks = true,
    linkcolor = {mycolor1},
    citecolor ={mycolor1}
}

\DeclareMathOperator*{\argmin}{argmin}
\DeclareMathOperator*{\argmax}{argmax}

\title{\bf MF-SCBO : Multi-fidelity Scalable Constrained
Bayesian Optimization}
\date{\today}

\author[1]{Lucas Palazzolo\orcidlink{0009-0002-4188-014X}}
\author[2]{Mickaël Binois\orcidlink{0000-0002-7225-1680}}
\author[1]{Laetitia Giraldi\orcidlink{0000-0003-2684-0203}}

\affil[1]{\small Université Côte d'Azur, Inria, Calisto team, Sophia Antipolis, France}
\affil[2]{\small Université Côte d'Azur, Inria, Acumes team, CNRS, LJAD, Sophia Antipolis, France}
\begin{document}

\maketitle

\begin{abstract}
Many real-world optimization problems rely on expensive simulations or experiments, making the efficient use of available data essential. Multi-fidelity optimization of high-dimensional black-box functions subject to black-box constraints is increasingly relevant as the cost of objective evaluations continues to rise in applications such as machine learning, engineering, and control. To our knowledge, no existing method simultaneously addresses high-dimensionality, black-box constraints, an arbitrary number of fidelity levels, and non-nested sampling. In this work, we extend the Scalable Constrained Bayesian Optimization method to the multi-fidelity setting, resulting in the MF-SCBO method. The proposed approach is evaluated on standard benchmark functions as well as challenging problems. The experimental results demonstrate that MF-SCBO generally achieves better convergence than both the single-fidelity SCBO and the other multi-fidelity method considered in these high-dimensional and constrained settings.
\end{abstract}

\section{Introduction}

Many problems in science and engineering consider the optimization of a black-box objective function that may be noisy, computationally expensive, or lack accessible gradient information. Bayesian optimization (BO) has been developed to address such settings \cite{Garnett2023,pourmohamad2021bayesian}. It has been successfully applied in a wide range of domains, including aerospace engineering \cite{Lam2018}, hyperparameter tuning in machine learning \cite{snoek2012}, robotics applications such as bipedal locomotion \cite{Rai2019}, multi-robot coordination \cite{Nogueira2016}, cybersecurity \cite{Deepthi2022}, and microswimmer control \cite{palazzolo2026}. The objective is to locate a global optimum while limiting the number of costly function evaluations, typically by using Gaussian process (GP) surrogates \cite{Rasmussen2006,Gramacy2020}.\\

As the evaluation costs of the objective increase, standard optimization approaches become unusable. This motivates the use of multi-fidelity optimization techniques \cite{do2025,Brevault2020}, where cheaper, lower-fidelity approximations of the objective function are available. These approximations are less accurate but can significantly reduce the computational cost for designs outside the area of interest. Thus the goal is to exploit them to guide the search and reduce the overall optimization budget compared to approaches relying only on high-fidelity evaluations.\\

Several approaches have been proposed for constructing multi-fidelity surrogate models. Linear models express a higher-fidelity surrogate as a sum of lower-fidelity surrogates. Methods for selecting such linear models include the autoregressive model of Kennedy and O'Hagan (KOH) \cite{kennedy2000}, hierarchical kriging \cite{han2012}, and recursive models \cite{legratiet2014}. Non-linear relationships between fidelities can be captured by compositions of GPs \cite{Brevault2020}. Non-linear autoregressive models allow correlations between fidelity levels to be learned, represented as a composition of GPs forming Deep Gaussian Processes (DGPs) \cite{perdikaris2017}, where each hidden layer is a GP \cite{damianou2013}. Multi-fidelity Deep Gaussian Processes (MF-DGPs) assign each fidelity level to a hidden layer \cite{cutajar2019}. The posterior in MF-DGPs is intractable, requiring approximation techniques that can be costly and may reduce accuracy. The methods described above model surrogates in the input space. An alternative is to treat fidelity as an additional input variable, allowing surrogates to capture variation along the fidelity dimension. Fidelity can be modeled as a continuous variable using \emph{continuous approximations} \cite{kandasamy2017}, or as a discrete variable with non-continuous covariance functions \cite{roustant2020}. Another aspect concerns the distinction between nested and non-nested sampling designs, depending on whether the set of evaluation points at a given fidelity level is included in the set of points at a lower fidelity level \cite{Sacher2021,baillie2025}. Extensions to non-ordered information sources are discussed by \cite{Poloczek2017}. These approaches are also related to multi-task and transfer learning, where information is shared across related tasks or information sources \cite{Swersky2013,tighineanu2022transfer}. To take into account the choice of fidelity when evaluating new points, acquisition functions have been extended to depend not only on the input variables but also on fidelity. This allows the selection process to consider both the potential error reduction and computational cost associated with lower-fidelity evaluations. In this setting, the acquisition function is jointly maximized over input and fidelity \cite{alexandrov1998,pang2017}. An alternative approach is sequential fidelity selection, where the new input point is chosen first and the corresponding fidelity is selected afterwards. In this case, the choice of the input point is independent of the fidelity level \cite{pang2017,kandasamy2017,tran2020}.\\

Despite extensive work in this area, it remains challenging for existing methods to simultaneously address multi-fidelity Bayesian optimization with an arbitrary number of fidelity levels, high-dimensionality, multiple constraints, and non-nested sampling settings (no inclusion is assumed between datasets at different fidelity levels). Combinations between Bayesian optimization and trust-region methods as in \cite{Eriksson2019,Diouane2021} have been successful in improving the scalability with respect to the input dimension \cite{Santoni2024}. The handling of constraints was handled with the Scalable Constrained Bayesian Optimization (SCBO) method \cite{eriksson_scalable_2021}, and in this work, we further extend it to the multi-fidelity setting. SCBO has proven to be effective for large-scale constrained problems through the use of trust regions, transformation of constraints, and Thompson sampling. For handling multi-fidelity, an additional question is on the choice of the trust region center when the number of high fidelity observations is small. Two variants are introduced here: \textit{evaluated} and \textit{predicted}. In the first, the trust-region center is determined from high-fidelity evaluations only. In the second, it is defined using high-fidelity predictions over all available data. Still, relying directly on the highest-fidelity posterior mean can lead to an unstable trust-region center identification. To mitigate this effect, we introduce a leave-one-out (LOO) criterion to assess the predictive accuracy. Constraints are assumed to be independent of the fidelity level, i.e., a single constraint function is shared across fidelities.\\

The proposed method is compared with the standard SCBO and the Multi-Fidelity Max-value Entropy Search (MF-MES) \cite{takeno2020}, when computationally feasible, on a range of benchmark problems. These include synthetic functions as well as more realistic and challenging applications, whose input dimension ranges from $6$ to $100$.\\

The paper is organized as follows. \Cref{sec:preliminaries} provides a brief overview of Gaussian processes and Bayesian optimization. The SCBO method and the multi-fidelity framework are also presented. The proposed MF-SCBO method is introduced in \Cref{sec:mfscbo}. Experimental settings and results are presented in \Cref{sec:experiments}. Finally, \Cref{sec:conclusion} summarizes and concludes the paper.

\section{Preliminaries}\label{sec:preliminaries}

\subsection{Gaussian Processes and Bayesian Optimization}

A Gaussian process is a stochastic process, a family of random variables, defined over a space $\Omega$. The law of a GP is entirely determined by its mean function $m : \Omega \to \mathbb{R}$ and its positive semi-definite covariance function $k :\Omega^2\to \mathbb{R}$. Let $f : \Omega \to \mathbb{R}$ be the function we wish to model. We assume that the prior distribution is a GP, i.e $\hat{f} \sim \mathcal{GP}(m,k)$ and $\hat{f}(\boldsymbol{x})\sim \mathcal{N}(m(\boldsymbol{x}), k(\boldsymbol{x}, \boldsymbol{x}))$ for all $\boldsymbol{x}$ in $\Omega$. For notational convenience, we will typically assume a zero mean function i.e, $m\equiv0$ (or a small value for numerical stability \cite{gramacy2011}). This assumption is not restrictive, since the posterior mean of the process is not constrained to be zero. Let $\mathcal{D}_n :=\{(\boldsymbol{x}_i, y_i)\}_{i=1}^n$ be a finite set of $n \in \mathbb{N}$ realizations satisfying $y_i = f(\boldsymbol{x}_i)+\varepsilon_i$ where the noise $\varepsilon_i$ follows $\mathcal{N}(0,\sigma_{\varepsilon}^2)$.  In this work, we restrict ourselves to the noise-free setting, i.e., $\sigma_{\varepsilon} = 0$. Denote by $X_n$ the matrix where the $i$-th row is associated to $\boldsymbol{x}_i$ and by $\boldsymbol{y}_n$ the vector of $y_i$. Then, $\hat{f}\mid \mathcal{D}_n$, called the posterior process is also a GP with mean function $\mu$ and covariance $\Sigma$ defined by 
\begin{align*}
    \mu(\boldsymbol{x}) &= k(\boldsymbol{x}, X_n)(K_n+\sigma_{\varepsilon}^2 I_n)^{-1} \boldsymbol{y}_n ,\\
    \Sigma(\boldsymbol{x}) &= k(\boldsymbol{x},\boldsymbol{x}) - k(\boldsymbol{x}, X_n)(K_n + \sigma_{\varepsilon}^2 I_n)^{-1} k(X_n, \boldsymbol{x}).
\end{align*}
Here, $K_n \in \mathbb{R}^{n \times n}$ denotes the covariance matrix with entries $(K_n)_{ij} = k(\boldsymbol{x}_i,\boldsymbol{x}_j)$ for $\boldsymbol{x}_i,\boldsymbol{x}_j \in X_n$. Gaussian processes are parameterized by several unknown quantities, depending on the assumptions made such as the presence of observation noise or the choice of covariance kernel. These parameters are referred to as \textit{hyperparameters}, typically selected via maximum likelihood estimation. The selection of hyperparameters plays a crucial role in building a reliable model while poorly chosen hyperparameters can lead to suboptimal models, see for instance guidelines by \cite{Gu2018,Marrel2024a}. For further details about GP fitting, see, e.g., \cite{Rasmussen2006, Gramacy2020}.\\

\noindent The optimization is performed under a set of inequality constraints defined by a function $\boldsymbol{c} : \Omega \to \mathbb{R}^m$, with $m \in \mathbb{N}^*$, such that constraints are written as $\boldsymbol{c}(\boldsymbol{x}) \leq 0$. The problem can be stated as:
\begin{equation}\label{eq:problem_setup}
    \inf_{\boldsymbol{x} \in \Omega} f(\boldsymbol{x})
    \quad \text{s.t.} \quad \boldsymbol{c}(\boldsymbol{x}) \leq 0 .
\end{equation}
where $\Omega$ denotes the search space, typically a bounded hypercube.  The core idea in BO is to use the surrogate $\hat{f}$ to guide the selection of new evaluation points, balancing exploration of poorly known regions of the input space with exploitation of the promising regions. This is generally achieved via an acquisition function $a:\Omega \to \mathbb{R} $, which quantifies the utility of sampling at a given location. The next evaluation point is selected by maximizing this acquisition function:
\begin{equation*}
\boldsymbol{x_{\text{next}}} \in \argmax_{\boldsymbol{x} \in\Omega} a(\boldsymbol{x}).
\end{equation*}
After evaluating the true objective function at $ \boldsymbol{x_{\text{next}}} $, the dataset is updated, and the GP model is refitted. This iterative procedure continues until a convergence criterion is met or a maximum number of iterations is reached. A high-level overview of the algorithm is presented in \Cref{alg:bo}.
\begin{algorithm}[htpb]
\DontPrintSemicolon
\SetKwInOut{Input}{Inputs}
\SetKwInOut{Output}{Output}
\Input{
    Objective function $f$, domain $\Omega$, acquisition function $a$, stopping criterion, maximum number of iterations $N$
}
\Output{Estimated minimizer $\boldsymbol{x}^*$}
\BlankLine

Initialize dataset: 
$\mathcal{D}_n = \{(\boldsymbol{x}_i, y_i) \mid y_i = f(\boldsymbol{x}_i)+\varepsilon_i, \ \boldsymbol{x}_i \in\Omega \} $

\While{stopping criterion not met and $n < N$}{
    Build or update the GP surrogate model $\hat{f}$ using $\mathcal{D}_n$

    Select the next evaluation point by maximizing the acquisition function:
    \begin{equation*}
    \boldsymbol{x_{\text{next}}} \in \argmax_{\boldsymbol{x} \in\Omega} a(\boldsymbol{x})
    \end{equation*}

    Evaluate the objective: $ y_{\text{next}} = f(\boldsymbol{x_{\text{next}}}) $

    Augment the dataset:
    \begin{equation*}
  \mathcal{D}_{n+1} =\mathcal{D}_n \cup \{(\boldsymbol{x_{\text{next}}}, y_{\text{next}})\} 
     \end{equation*}

    $ n \gets n + 1 $
}

\Return{$ \boldsymbol{x}^* \in \argmin_{\boldsymbol{x}\in \mathcal{X}_n} f(\boldsymbol{x})$ }
\caption{Bayesian Optimization}
\label{alg:bo}
\end{algorithm}
This procedure is iterated until a stopping criterion is met or a maximum number of iterations is reached. A high-level description of the algorithm is given in \Cref{alg:bo}. Among many options, summarized for instance in \cite{Garnett2023}, common acquisition functions include Expected Improvement (EI) \cite{Mockus1975}, Upper Confidence Bound (UCB) \cite{Srinivas2009}, or those based on the entropy of the solution like \cite{Villemonteix2009}. Extensions to batch settings, such as $q$-EI \cite{Ginsbourger2010a}, have also been proposed to allow multiple evaluations per iteration. Batch strategies are particularly useful when parallel evaluations are available, as they reduce wall-clock optimization time. More details are given in \cite{Garnett2023}.

\subsection{High dimension and constraints}\label{sec:scbo}

In high-dimensional settings, not only building a global surrogate model is difficult but the optimization of acquisition functions can also cause some problems, see e.g., \cite{Binois2022}. Their gradients tend to vanish as the dimension increases \cite{papenmeier2025}, because vast unexplored regions can cause acquisition functions to become nearly flat. Trust-region methods address this issue by restricting the search to local subregions. A notable example is the Trust Region Bayesian optimization (TuRBO) \cite{Eriksson2019} method, which also replaces standard acquisition functions with Thompson sampling \cite{Thompson1933}, avoiding the acquisition function optimization to focus on the boundary of the Trust Region (TR).  A disturbance probability, Random Axis-Aligned Subspace Perturbation (RAASP) as introduced in \cite{rashidi2024,eriksson_scalable_2021}, has been incorporated to mitigate vanishing gradients and the edge effect. The edge effect refers to the phenomenon where points in high-dimensional spaces tend to lie near the boundaries of the search space. This effect can trap the algorithm at the boundaries of the hypercube.\\

\noindent An extension of these methods, Scalable Constrained Bayesian optimization (SCBO), incorporates constraints directly into the optimization process \cite{eriksson_scalable_2021}. Its principle is as follows. Starting from an initial set of evaluated points, the feasible set is defined as the set of points satisfying all constraints. If feasible points are available, the center of the TR is chosen as the best feasible solution in terms of the objective; otherwise, the point with the smallest overall constraint violation is selected. A local trust region, typically a hypercube centered at this reference point, is then defined, with its size controlling the extent of exploration.\\

\noindent Within this region, candidate points are generated using a Thompson sampling strategy. Samples are drawn from the posterior distributions of both the objective and constraints over a discretized set. If feasibility is reached, the next point is selected as the best feasible candidate; otherwise, the point with minimal constraint violation is chosen. The selected candidates are evaluated using the true objective and constraint functions, and the resulting observations are used to update the surrogate models.\\

\noindent The trust region is updated adaptively based on observed progress. It is expanded when improvements are observed, and contracted otherwise, focusing the search locally when progress stagnates. The TR center is updated to the best available point, and the procedure is repeated until the evaluation budget is exhausted or the region becomes too small.

\subsection{Modeling multi-fidelity}

Our objective is to minimize a function $f$ defined on a domain $\Omega$, which is assumed to be very costly to evaluate (in terms of computation time and/or memory). The problem is defined in \eqref{eq:problem_setup}.\\

\noindent Let $\{f^{(s)}\}_{s=0}^{S}$ be a family of functions of increasing fidelity, and let $\{\eta^{(s)}\}_{s=0}^{S} \subset \mathbb{R}_{+}$ denote their associated evaluation costs. The index $s$ represents the fidelity level, with $s=0$ corresponding to the lowest fidelity and $s=S$ to the highest one, so that $f^{(S)} = f$. For all $s \in \{0,\ldots,S-1\}$, the following assumptions are made, see, e.g., \cite{kandasamy2019}:
\begin{enumerate}
    \item the approximation error satisfies
    \begin{equation*}
        \| f^{(s)} - f \|_{L^{\infty}(\mathcal{X})} \leq \varepsilon^{(s)},
        \quad \text{with } \varepsilon^{(s+1)} \leq \varepsilon^{(s)}
        \text{ and } \varepsilon^{(S)} = 0,     
    \end{equation*}
    \item the evaluation cost increases with the fidelity level:
    \begin{equation*}
        \eta^{(s+1)} \geq \eta^{(s)}.     
    \end{equation*}
\end{enumerate}
The first assumption states that higher-fidelity models provide more accurate approximations of the target function $f$, while the second reflects their increased computational cost. The objective of \textit{multi-fidelity optimization} is to exploit low-fidelity models to reduce the number of evaluations of the high-fidelity function while preserving optimization accuracy.\\

\noindent Several approaches exist to combine multiple fidelities with GP models. Here, we adopt the auto-regressive recursive formulation introduced by \cite{legratiet2014}. Each surrogate $\hat{f}^{(s)}$ is a GP with mean $\mu^{(s)}$ and covariance $\Sigma^{(s)}$. The lowest-fidelity model $\hat{f}^{(0)}$ is defined directly, and for each $s = 0, \ldots, S-1$, we introduce an independent GP $\hat{\delta}^{(s)}$ representing the error between two successive fidelity levels. The models satisfy the recursive relation
\begin{equation*}
    \hat{f}^{(s+1)} = \rho^{(s)} \hat{f}^{(s)} + \hat{\delta}^{(s)},
\end{equation*}
where $\rho^{(s)} \in \mathbb{R}$ is a scaling parameter learned by maximizing the marginal likelihood of $\hat{\delta}^{(s)}$ along with the other GP hyperparameters. This parameter controls the correlation between fidelity levels: a small value of $\rho^{(s)}$ indicates weak correlation and implies that $\hat{f}^{(s+1)}$ relies mainly on the correction $\hat{\delta}^{(s)}$ (or just this fidelity level's values when $\rho^{(s)} = 0$), whereas a large value indicates a difference of output scale between fidelities. In practice, we constrained $\rho^{(s)}$ to the interval $[0, 2]$ to ensure numerical stability. Then, we obtain for $s=0,\ldots, S-1$ :
\begin{align*}
    \mu^{(s+1)}(\boldsymbol{x}) &= \rho^{(s)}\mu^{(s)}(\boldsymbol{x})+k_{\delta^{(s)}}(\boldsymbol{x},X^{(s+1)})K_{\delta^{(s)}}^{-1}(\boldsymbol{y}^{(s+1)}-\rho^{(s)}\boldsymbol{\mu}^{(s)}(X^{(s+1)})),\\
    \Sigma^{(s+1)}(\boldsymbol{x})&=(\rho^{(s)})^2\Sigma^{(s)}(\boldsymbol{x})+k_{\delta^{(s)}}(\boldsymbol{x},\boldsymbol{x})-k_{\delta^{(s)}}(\boldsymbol{x},X^{(s+1)})K_{\delta^{(s)}}^{-1}k_{\delta^{(s)}}(X^{(s+1)},\boldsymbol{x}).
\end{align*}

\begin{figure}[htpb]
    \centering
    \includegraphics[width=0.6\linewidth]{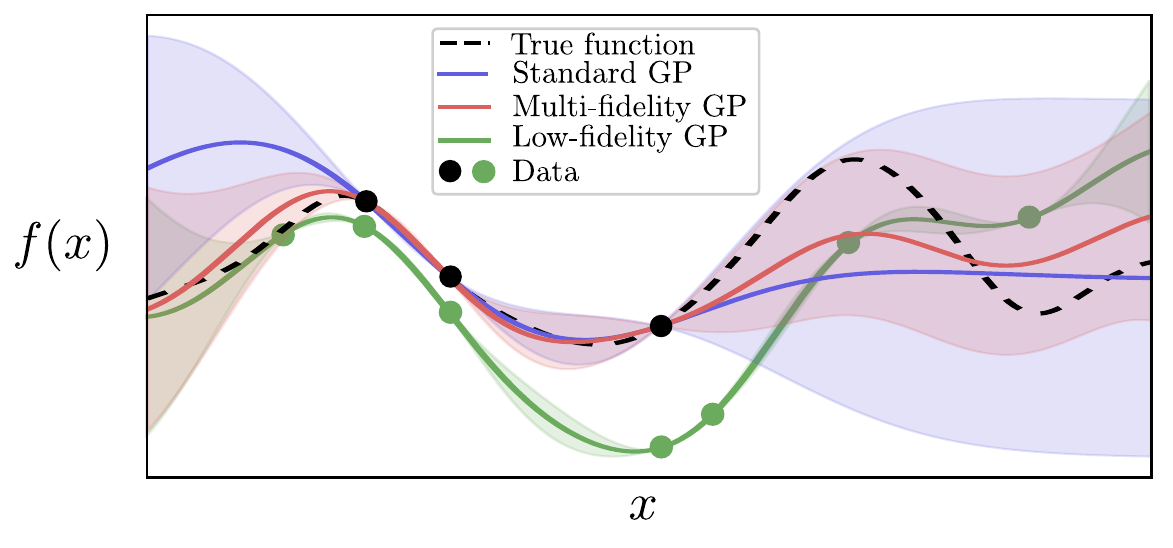}
\caption{Multi-fidelity Gaussian process regression on a toy function. Multi-fidelity GP (red) constructed from a small number of high-fidelity data points and a larger set of low-fidelity data points (red). The nested case is considered, where the high-fidelity samples are a subset of the low-fidelity points. The standard GP built using only high-fidelity data is displayed in blue. The $95\%$ confidence intervals predicted by the GPs are shown as shaded regions.}
    \label{fig:mfreg}
\end{figure}

\noindent Let $(\mathcal{X}^{(s)}, \mathcal{Y}^{(s)})$ denote the set of training points used to construct the surrogate $\hat{f}^{(s)}$ and $(\Delta\mathcal{X}^{(s)},\Delta\mathcal{Y}^{(s)})$ denote the set of training points used to construct the surrogate $\hat{\delta}^{(s)}$. Two configurations are commonly considered. In the \textit{nested} case, the training sets satisfy
\begin{equation}\label{eq:nested}
    \mathcal{X}^{(s+1)} \subset \mathcal{X}^{(s)}, \quad s = 0, \ldots, S-1,
\end{equation}
meaning that higher-fidelity data are available only at a subset of the lower-fidelity data. This structure simplifies the learning of the error processes. In the \textit{non-nested} case, no such inclusion holds, which complicates the construction and training of the error models $\hat{\delta}^{(s)} = \hat{f}^{(s+1)} - \rho^{(s)} \hat{f}^{(s)}.$ To distinguish between these two configurations, we use the notations
$\Delta^{\mathrm{N}}\cdot$ and $\Delta^{\mathrm{NN}}\cdot$ to denote the nested
and non-nested cases, respectively. \Cref{fig:mfreg} illustrates multi-fidelity Gaussian process regression on a toy function in the case of two fidelity levels.\\

\noindent A simple extension to the non-nested case is to replace missing observations by their predictive means as in \cite{Sacher2021}. For each fidelity level $s \in \{0,\ldots,S-1\}$, we distinguish between the evaluation points shared by fidelities $s$ and $s+1$ and those that are unique to fidelity $s+1$. We define the set of common points by $\mathcal{X}_{\mathrm{in}}^{(s+1)}$ and the set of points exclusive to fidelity $s+1$ by $\mathcal{X}_{\mathrm{out}}^{(s+1)}$, given by
\begin{equation*}
\mathcal{X}_{\text{in}}^{(s+1)} := \mathcal{X}^{(s+1)} \cap \mathcal{X}^{(s)}, 
\qquad
\mathcal{X}_{\text{out}}^{(s+1)} := \mathcal{X}^{(s+1)} \setminus \mathcal{X}^{(s)}.
\end{equation*}
When common observations are unavailable, we rely on the predictions $\mu^{(s)}$. We therefore divide our set $\mathcal{X}^{(s+1)}$ into two: the points common to both fidelities $\mathcal{X}_{\text{in}}^{(s+1)}$ and the points only in fidelity $s+1$, $\mathcal{X}_{\text{out}}^{(s+1)}$. The training set for $\hat{\delta}^{(s)}$ is
 \begin{equation*}
(\Delta\mathcal{X}^{(s)}, \Delta\mathcal{Y}^{(s)})
= \left\{ (\boldsymbol{x}_i^{(s+1)}, \Delta^{\mathrm{NN}} y_i^{(s)}) \right\}_{i=1}^{n_{s+1}},
\end{equation*}
with
\begin{equation}\label{eq:deltanonnested}
\Delta^{\mathrm{NN}} y_i^{(s)} :=
\begin{cases}
\Delta^{\mathrm{N}} y_i^{(s)}, 
& \quad\text{if }\boldsymbol{x}_i^{(s+1)} \in \mathcal{X}_{\text{in}}^{(s+1)}, \\
y_i^{(s+1)} - \rho^{(s)} \mu^{(s)}(\boldsymbol{x}_i^{(s+1)}),
& \quad\text{if }\boldsymbol{x}_i^{(s+1)} \in \mathcal{X}_{\text{out}}^{(s+1)}.
\end{cases}
\end{equation}
with \begin{equation}\label{eq:deltanested}
\Delta^{\text{N}} y_i^{(s)} := y_i^{(s+1)} - \rho^{(s)} y_i^{(s)},
\qquad
y_i^{(s)} = f^{(s)}(\boldsymbol{x}_i^{(s+1)}) = f^{(s)}(\boldsymbol{x}_{\sigma(i)}^{(s)}),
\end{equation}
where $\sigma$ denotes a mapping that associates the common points between $\mathcal{X}^{(s+1)}$ and $\mathcal{X}^{(s)}$.\\

\noindent Other approaches can be considered to account for the additional uncertainty via an expectation-maximization scheme \cite{baillie2025}.

\section{Multi-fidelity Bayesian Optimization for High-Dimensional Constrained Problems}\label{sec:mfscbo}

\begin{figure}[h]
    \centering
    \includegraphics[width=1\linewidth]{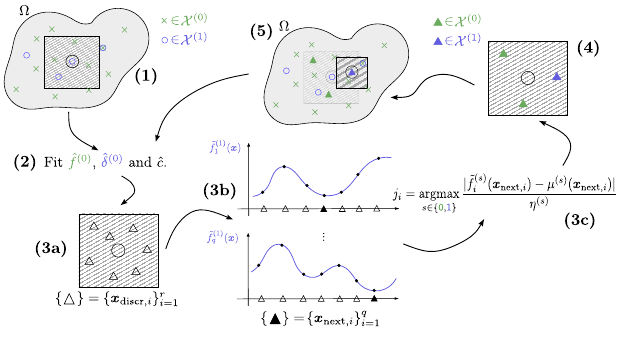}
\caption{MF-SCBO representation for two fidelity levels (green for the low fidelity and blue for the high fidelity): \textbf{(1)} Initial points are taken from $\Omega$ (represented by crosses for the low fidelity and circle for the high fidelity), and the trust region is defined (grid with central black circle). \textbf{(2)} The different Gaussian process models are fitted. \textbf{(3a)} $r$ random points (represented by triangles) are sampled in the trust region. \textbf{(3b)} A batch of $q$ realizations is computed and selected via Thompson sampling on these $r$ designs. For each realization, the $x$ axis corresponds to the $\boldsymbol{x}$ points in $\Omega$ and is discretized by the $\{\boldsymbol{x}_{\text{discr},i}\}_{i=1}^r$ (the triangles) chosen previously. The $y$ axis shows the realizations. For simplicity, a single graph is shown for the objective and the constraints. For each realization, we keep the best point (completely filled triangle). \textbf{(3c)} For the $q$ best points $\{\boldsymbol{x}_{\text{next},i}\}_{i=1}^q$, the fidelity level is choose by computing \eqref{eq:assignfid}.
\textbf{(4)} The real value of the objectives and the constraints are computed and added to the previous observations. \textbf{(5)} The trust region is readjusted by modifying its center and radius.}
    \label{fig:mfscbo}
\end{figure}

To solve constrained optimization problems involving many variables while exploiting multi-fidelity, we propose the MF-SCBO algorithm, extending the SCBO algorithm (see \Cref{sec:scbo}), originally introduced in \cite{eriksson_scalable_2021}. The use of \textit{Thompson Sampling} (TS) provides a computationally efficient alternative to standard acquisition functions, which can be difficult to optimize, especially in high-dimensional settings or in the presence of multiple fidelity levels. To the best of our knowledge, TS has not been used in the context of multi-fidelity Bayesian optimization. Two main challenges arise: (i) defining a TS strategy that accounts for multiple fidelity levels, and (ii) integrating such a strategy within a trust-region framework. We consider two strategies for updating the trust region: the \textit{evaluation} method and the \textit{prediction} method. In the evaluation method, the trust region is updated exclusively based on high-fidelity evaluations; consequently, the update is performed only once a full batch of high-fidelity points has been obtained. The prediction method updates the trust region using points from all fidelities, relying on predictions made by the high-fidelity surrogate model. The \textit{evaluated} method is illustrated for two fidelities in \Cref{fig:mfscbo}.\\

\noindent In the following, we introduce the notation used throughout the method. The total number of observations is  $n := \sum_{s=0}^S n_s$. We assume that the constraints are independent of the fidelity level. The set of constraint observations is therefore
\begin{equation*}
\mathcal{C}_n := \left\{ (\boldsymbol{x}_i, \boldsymbol{c}_i) \mid 
\boldsymbol{c}_i = \boldsymbol{c}(\boldsymbol{x}_i),\; \boldsymbol{x}_i \in \bigcap_{s=0}^S \mathcal{X}^{(s)} \right\}.
\end{equation*}
For each fidelity level $s \in \{0,\ldots,S\}$, we consider a set of $n_s$ observations
\begin{equation*}
\mathcal{D}_{n_s}^{(s)} := \left\{ (\boldsymbol{x}_i^{(s)}, y_i^{(s)})\mid y_i^{(s)} = f^{(s)}(\boldsymbol{x}_i^{(s)}) \right\}_{i=0}^{n_s}.
\end{equation*}

\subsection{Trust Region Management}

The feasible set is defined as
\begin{equation}\label{eq:feasiblesetmfscbo}
F_{\boldsymbol{c}} := \left\{ \boldsymbol{x} \mid 
\boldsymbol{c}(\boldsymbol{x})\leq 0,\;
\boldsymbol{x} \in \bigcap_{s=0}^S \mathcal{X}^{(s)} \right\}.
\end{equation}
The center of the trust region depends on the update strategy : evaluation-based approach or predicted-based approach. 

\paragraph{Evaluation-based approach.}

In the evaluation-based approach, only high-fidelity evaluations are considered, leading to the center
\begin{equation}\label{eq:x_eval}
    \boldsymbol{x}_{\text{eval}}^{*} :=\begin{cases}
        \displaystyle{\argmin_{\boldsymbol{x}\in F_{\boldsymbol{c}}\cap\mathcal{X}^{(S)}}} f^{(S)}(\boldsymbol{x}), \quad &\text{if } F_{\boldsymbol{c}}\cap\mathcal{X}^{(S)}\neq \emptyset\\
        \displaystyle{\argmin_{\boldsymbol{x}\in\bigcap_{s=0}^{S}\mathcal{X}^{(s)}}}\sum_{i=1}^m\max\{c^j(\boldsymbol{x}),0\}, \quad &\text{else}.
    \end{cases}
\end{equation}
This means that if $F_{\boldsymbol{c}}\cap \mathcal{X}^{(S)}$is nonempty, the center is chosen as the point minimizing the objective function. Otherwise, it is defined as the point minimizing the maximum constraint violation. The trust-region size is initialized with a length $L_{\text{init}}$, and the corresponding region is given by $\mathcal{B}(\boldsymbol{x}^*_{\text{eval}}, L_{\text{init}})$. In the single-fidelity SCBO setting, the same principle is applied using the evaluation-based strategy with only the highest fidelity level considered. In this case, the trust region is updated only after $q$ points of the highest fidelity have been evaluated. If a new candidate is better than the actual one, the center of the region is shifted, corresponding to $\boldsymbol{x}_{\mathrm{eval}}^{*}$ in \eqref{eq:x_eval}.

\paragraph{Prediction-based approach.}
In the prediction-based approach, the center is obtained by considering high-fidelity prediction over the whole data set and given by
\begin{equation}\label{eq:x_pred}
    \boldsymbol{x}_{\text{pred}}^{*} :=\begin{cases}
        \displaystyle{\argmin_{\boldsymbol{x}\in F_{\boldsymbol{c}}}}~~ \mu^{(S)}(\boldsymbol{x}) \quad &\text{if } F_{\boldsymbol{c}}\neq \emptyset\\
        \displaystyle{\argmin_{\boldsymbol{x}\in\bigcap_{s=0}^{S}\mathcal{X}^{(s)}}}\sum_{i=1}^m\max\{c_j(\boldsymbol{x}),0\} \quad &\text{else}
    \end{cases}
\end{equation}
This means that if $F_{\boldsymbol{c}}$ is nonempty, the center is chosen as the point minimizing the high-fidelity prediction. Otherwise, it is defined as the point minimizing the maximum constraint violation. The trust-region size is initialized with a length $L_{\text{init}}$, and the corresponding region is given by $\mathcal{B}(\boldsymbol{x}^*_{\text{pred}}, L_{\text{init}})$. In this case, the trust region is adapted based on the highest-fidelity predictions of all points, regardless of their actual fidelity level. Since additional points are incorporated into the fitting of the Gaussian processes (GPs), the value of the current best point is updated according to the highest-fidelity mean prediction, denoted by $\mu^{(S)}$. This updated best point is then compared with the predictions associated with the $q$ selected points. If a better point is identified among the newly selected points, a leave-one-out (LOO) error criterion is used to assess the reliability of its highest-fidelity prediction. Let $s^*$ denote the fidelity level of this point and $i^*$ its index in the set $\mathcal{X}^{(s^*)}$. The LOO procedure consists of removing a given point from the training set and evaluating the corresponding GP prediction at the removed point. It provides a criterion for assessing whether the new candidate point, identified from the highest-fidelity prediction, is sufficiently well predicted. If the prediction is considered unreliable, the point is evaluated at the highest fidelity level to prevent an inaccurate prediction from becoming the center of the trust region. For a single fidelity GP, the LOO mean and variance at the removed point are given by (see, e.g., \cite{Dubrule1983})
\begin{equation}\label{eq:classical_loo}
\mu_{-i}(\boldsymbol{x}_{i})
= y_{i} - \frac{(K^{-1}\boldsymbol{y})_{i}}{(K^{-1})_{ii}},
\qquad
\sigma_{-i}^{2}(\boldsymbol{x}_i)
= \frac{1}{(K^{-1})_{ii}},
\end{equation}
where $i$ denotes the index of the point removed from the training data. Since the multifidelity GP formulation is sequential, the LOO mean prediction at fidelity $s^*$ can be written as
\begin{equation}\label{eq:loo_mean}
\mu^{(s^*)}_{-i^*}
\left(\boldsymbol{x}^{(s^*)}_{i^*}\right)
=
\rho^{(s^*-1)}
\mu_{-i^*}^{(s^*-1)}
\left(\boldsymbol{x}^{(s^*)}_{i^*}\right)
+
\mu_{\delta^{(s^*-1}),-i^*}
\left(\boldsymbol{x}^{(s^*)}_{i^*}\right).
\end{equation}
Let
\begin{equation}
r_{i^*}^{(s^*-1)}
:=
y_{i^*}^{s^*}
-
\rho^{s^*-1}
\mu^{(s^*-1)}
\left(\boldsymbol{x}^{(s^*)}_{i^*}\right).
\end{equation}
Using \eqref{eq:classical_loo} and \eqref{eq:deltanonnested}, the LOO prediction of the discrepancy GP is
\begin{equation}\label{eq:loo_mean_delta}
\mu_{\delta^{s^*-1},-i^*}
\left(\boldsymbol{x}^{(s^*)}_{i^*}\right)
=
r_{i^*}^{(s^*-1)}
-
\frac{
\left(K_{\delta^{(s^*-1)}}^{-1}
\boldsymbol{r}^{(s^*-1)}\right)_{i^*}
}{
\left(K_{\delta^{(s^*-1)}}^{-1}\right)_{i^*i^*}
}.
\end{equation}
Moreover, the lower-fidelity GP is unaffected by the removal of this point, since the point belongs to the fidelity level $s^*$:
\begin{equation}\label{eq:loo_mean_lower_fid}
\mu_{-i^*}^{(s^*-1)}
\left(\boldsymbol{x}^{(s^*)}_{i^*}\right)
=
\mu^{(s^*-1)}
\left(\boldsymbol{x}^{(s^*)}_{i^*}\right).
\end{equation}
Substituting \eqref{eq:loo_mean_delta} and \eqref{eq:loo_mean_lower_fid} into \eqref{eq:loo_mean}, and applying the same reasoning to the variance, yields
\begin{align}\label{eq:loo_mean_var_simp}
\mu^{(s^*)}_{-i^*}
\left(\boldsymbol{x}^{(s^*)}_{i^*}\right)
&=
y_{i^*}^{(s^*)}
-
\frac{
\left(K_{\delta^{(s^*-1)}}^{-1}
\boldsymbol{r}^{(s^*-1)}\right)_{i^*}
}{
\left(K_{\delta^{(s^*-1})}^{-1}\right)_{i^*i^*}
},\\
\left(\sigma_{-i^*}^{(s^*)}\right)^2
\left(\boldsymbol{x}^{(s^*)}_{i^*}\right)
&=
\left(\rho^{(s^*-1)}\right)^2
\left(\sigma^{(s^*-1)}\right)^2
\left(\boldsymbol{x}^{(s^*)}_{i^*}\right)
+
\frac{1}{
\left(K_{\delta^{(s^*-1})}^{-1}\right)_{i^*i^*}
}.
\end{align}
The reliability of the highest-fidelity prediction is given by comparing the LOO error with the corresponding $95\%$ confidence interval. More precisely, the criterion is given by \begin{equation} \label{eq:loo_crit}\mathcal{C}_{\text{LOO}}=\underbrace{\left| y_{i^*}^{s^*} - \mu^{(s^*)}_{-i^*} \left(\boldsymbol{x}^{(s^*)}_{i^*}\right) \right|}_{\mathcal{E}_{\mathrm{LOO}}} - 1.96\, \sigma_{-i^*}^{(s^*)} \left(\boldsymbol{x}^{(s^*)}_{i^*}\right)\leq0, \end{equation} where $1.96\,\sigma_{-i^*}^{(s^*)}$ defines the $95\%$ confidence interval of the LOO prediction. If this condition is not satisfied, the candidate point is evaluated at the highest fidelity level in order to avoid using an unreliable prediction as the center of the trust region. Note that, thanks to the analytical form of the LOO prediction, no additional GP needs to be trained to compute the LOO error. Consequently, the proposed criterion introduces only a negligible computational cost.

\paragraph{Update of the length.}
The length of the trust region, $L^{t+1}$, is adjusted based on the number of successes or failures. Let $\tau_s$ and $\tau_f$ be the fixed success and failure rates, respectively, and $n_s$ and $n_f$ the number of successes and failures. Success is achieved when a better point than the center of the hypercube is obtained. The length of the hypercube is then adapted as follows:
\begin{equation*}
\text{if } n_s = \tau_s \text{ then } \begin{cases}
L^{t+1} = \min\{2L^{t}, L_{\text{max}}\},\\
n_s = 0,
\end{cases}
\text{ and }
\text{if } n_f = \tau_f \text{ then } \begin{cases}
L^{t+1} = L^{t}/2,\\
n_f = 0,
\end{cases}
\end{equation*}
with $L_{\text{max}}$ and $L_{\text{min}}$ denoting the maximum and minimum length of the hypercube. Otherwise, $L^{t+1}=L^{t}$. The trust region is then defined by $\mathcal{B}(\boldsymbol{x}^{t+1},L^{t+1})$.

\subsection{Thompson Sampling}

A batch of $q$ new points is selected using TS. First of all, $q$ samples from the posterior of cost and constraints are selected $\tilde{f}_i^{(s)} \sim \hat{f}^{(s)}\mid\mathcal{D}_{n_s}^{(s)}$ and $\tilde{\boldsymbol{c}}_i\sim \hat{\boldsymbol{c}}\mid \mathcal{C}_n$ for $i=1,\ldots, q$ and for $s=0,\ldots, S$. A candidate set of size $r$,  denoted by $\{\boldsymbol{x}_{\text{discr},i}\}_{i=1}^r$, is sampled to discretize the trust region. As \eqref{eq:feasiblesetmfscbo}, a feasible set $F_{\tilde{\boldsymbol{c}}_i}$ is defined based on $\tilde{\boldsymbol{c}}_i$ and  $\{\boldsymbol{x}_{\text{discr},i}\}_{i=1}^r$. For each posterior sample, the next evaluation point is selected as
\begin{equation*}
    \boldsymbol{x}_{\text{next},i} \in \begin{cases}
\displaystyle{\argmin_{\boldsymbol{x} \in F_{\tilde{\boldsymbol{c}}_i}} \tilde{f}_i^{(S)}(\boldsymbol{x})} \quad &\text{if } F_{\tilde{\boldsymbol{c}}_i} \neq \emptyset, \\
\displaystyle{\argmin_{\boldsymbol{x} \in \{\boldsymbol{x}_{\text{discr},l}\}_{l=1}^r} \sum_{j=1}^m \max\left\{\tilde{c}^j_i(\boldsymbol{x}), 0\right\}} \quad &\text{if } F_{\tilde{\boldsymbol{c}}_i} = \emptyset.
\end{cases}
\end{equation*}
The resulting batch is denoted by $\{\boldsymbol{x}_{\text{next},i}\}_{i=1}^q.$ In contrast to the classical single-fidelity SCBO method, a low-cost criterion is required to assign a fidelity level to each selected point. For each selected point, the fidelity level is chosen by
\begin{equation}\label{eq:assignfid}
j_i = \argmax_{s \in \{0,\ldots,S\}}
\frac{\left| \tilde{f}_i^{(s)}(\boldsymbol{x}_{\text{next},i}) - 
\mu^{(s)}(\boldsymbol{x}_{\text{next},i}) \right|}{\eta^{(s)}}, \qquad \text{for } i=1,\ldots,q.
\end{equation}
This criterion balances prediction uncertainty against evaluation cost, encouraging learning at lower fidelities whenever they remain insufficiently accurate to learn higher-fidelity models. A higher value of the criterion corresponds to larger prediction errors. The underlying objective is that, as long as the error of a low-fidelity model exceeds a threshold determined by the ratio of error to the cost of higher-fidelity evaluations, additional low-fidelity evaluations are prioritized. This strategy reduces the total number of expensive high-fidelity evaluations.\\

\noindent The MF-SCBO algorithm proceeds as follows and pseudo-code is presented in \Cref{alg:mfscbo}.

\begin{algorithm}[htbp]
\DontPrintSemicolon
\SetKwInOut{Input}{Inputs}
\SetKwInOut{Output}{Output}
\SetKwProg{Fn}{Function}{:}{end}
\SetKwFunction{UpdateGPs}{UpdateGPs}

\Input{
    Objective functions $\{f^{(s)}\}_{s=0}^{S}$, associated costs $\{\eta^{(s)}\}_{s=0}^{S}$,
    constraint function $\boldsymbol{c}$, domain $\Omega$, batch size $q$, stopping criterion
}
\Output{Estimated minimizer $\boldsymbol{x}^*$}
\BlankLine

\tcc{Subroutine: refit all surrogate models and the fidelity discrepancies}
\Fn{\UpdateGPs{}}{
    Update GP surrogate $\hat{c}$ using $\mathcal{C}_n$\;
    Update GP surrogate $\hat{f}^{(0)}$ using $\mathcal{D}_{n_0}^{(0)}$\;
    \For{$s=0$ \KwTo $S-1$}{
        \For{$i=1$ \KwTo $n_{s+1}$}{
                Compute $\Delta^{\mathrm{NN}} y_i^{(s)}$ \eqref{eq:deltanonnested}\;
        }
        Update discrepancy GP $\hat{\delta}^{(s)}$ and trust-region ratio $\rho^{(s)}$
        using $(\Delta\mathcal{X}^{(s)}, \Delta\mathcal{Y}^{(s)})$\;
    }
}

\BlankLine
\tcc{Initialization}
Initialize datasets $\mathcal{D}_{n_s}^{(s)}$ ($s=0,\dots,S$), $\mathcal{C}_n$, $\{\rho^{(s)}\}_{s=0}^{S} \gets \{1\}_{s=0}^{S}$\;
$\mathcal{X}_{\text{eval}}^{(S)}, \mathcal{Y}_{\text{eval}}^{(S)} \gets \emptyset, \emptyset$
    \tcp*[l]{buffer of highest-fidelity evaluations}
\UpdateGPs{}\;

\BlankLine
\While{stopping criterion not met}{
    Select a batch of $q$ candidate points $\{\boldsymbol{x}_{\text{next},i}\}_{i=1}^{q}$ via Thompson sampling\;

    \tcc{Assign a fidelity to each candidate and evaluate it}
    \For{$i=1$ \KwTo $q$}{
        \For{$s=0$ \KwTo $S$}{
            Sample $y_{\text{next},i}^{(s)} \sim \mathcal{N}\!\left(\mu^{(s)}(\boldsymbol{x}_{\text{next},i}), \Sigma^{(s)}(\boldsymbol{x}_{\text{next},i})\right)$\;
        }
        Select fidelity level $j$ \eqref{eq:assignfid}\;
        Evaluate objective $f^{(j)}$ and constraint $\boldsymbol{c}$ at $\boldsymbol{x}_{\text{next},i}$\;
        Augment $\mathcal{D}_{n_j}^{(j)}$; set $n_j \gets n_j + 1$\;

        \If{$j = S$}{
            $\mathcal{X}_{\text{eval}}^{(S)} \gets \mathcal{X}_{\text{eval}}^{(S)} \cup \{\boldsymbol{x}_{\text{next},i}\}$,
            $\mathcal{Y}_{\text{eval}}^{(S)} \gets \mathcal{Y}_{\text{eval}}^{(S)} \cup \{f^{(S)}(\boldsymbol{x}_{\text{next},i})\}$
                \tcp*[l]{evaluation case}
        }
    }

    \UpdateGPs{}\;

    \BlankLine
    \tcc{Update trust-region center}
    \If{evaluation case \textbf{and} $|\mathcal{X}_{\text{eval}}^{(S)}| = q$}{
        $y^{*}_{\text{eval}} \gets \min\{y^{*}_{\text{eval}}, \mathcal{Y}_{\text{eval}}^{(S)}\}$\;
        $\boldsymbol{x}_{\text{eval}}^{*} \gets \argmin_{\boldsymbol{x} \in \{\boldsymbol{x}_{\text{eval}}^{*}\} \cup \mathcal{X}_{\text{eval}}^{(S)}} f^{(S)}(\boldsymbol{x})$\;
        $\mathcal{X}_{\text{eval}}^{(S)}, \mathcal{Y}_{\text{eval}}^{(S)} \gets \emptyset, \emptyset$\;
    }

    \If{prediction case}{
        $y^{*}_{\text{pred}} \gets \mu^{(S)}(\boldsymbol{x}_{\text{pred}}^{*})$\;
        Compare current center $y^{*}_{\text{pred}}$ with highest-fidelity predictions $\{\mu^{(S)}(\boldsymbol{x}_{\text{next},i})\}_{i=1}^{q}$\;

        \tcc{Leave-one-out check}
        \If{a better point is found \textbf{and} $\mathcal{C}_{\text{LOO}} > 0$ \eqref{eq:loo_crit}}{
            $\boldsymbol{x}^{*} \gets \argmin_{\boldsymbol{x} \in \{\boldsymbol{x}_{\text{next},i}\}_{i=1}^{q}} \mu^{(S)}(\boldsymbol{x})$\;
            Evaluate $\boldsymbol{x}^{*}$ with the highest-fidelity function $f^{(S)}$\;
            Augment $\mathcal{D}_{n_S}^{(S)}$; set $n_S \gets n_S + 1$\;
        }

        $y^{*}_{\text{pred}} \gets \min\{y^{*}_{\text{pred}}, \mu^{(S)}(\boldsymbol{x}_{\text{next},1}), \dots, \mu^{(S)}(\boldsymbol{x}_{\text{next},q})\}$\;
        $\boldsymbol{x}_{\text{pred}}^{*} \gets \argmin_{\boldsymbol{x} \in \{\boldsymbol{x}_{\text{pred}}^{*}\} \cup \{\boldsymbol{x}_{\text{next},i}\}_{i=1}^{q}} \mu^{(S)}(\boldsymbol{x})$\;
    }

    Update trust-region length\;
}

\Return{$\boldsymbol{x}^*$}
\caption{MF-SCBO}
\label{alg:mfscbo}
\end{algorithm}

\newpage

\section{Experiments}\label{sec:experiments}

We compare MF-SCBO under two variants: (i) \emph{evaluated} and (ii) \emph{predicted}. The implementation has been developed using the \texttt{BoTorch} library \cite{balandat2020botorch} and is released as an open-source \texttt{Python} package: \texttt{MF-SCBO}\footnote{\url{https://github.com/Nemo-Virtual-Lab/MF-SCBO}}. These methods are benchmarked against Multi-Fidelity Max-value Entropy Search (MF-MES) \cite{takeno2020} and the single-fidelity SCBO method \cite{eriksson_scalable_2021}.\\

\subsection{Problem Setup}

Several multi-fidelity optimisation problems are considered. Five are based on standard synthetic benchmark functions, four of which include constraints : Hartmann, Borehole, Ackley and Rosenbrock \cite{letham2019,eriksson_scalable_2021,mainini2025}. In addition, four more realistic and challenging problems are studied as: solar thermal power plant simulator \cite{andres2025}; and airfoil shape optimization \cite{sobieczky1999,do2025} . The problem dimensions considered range from $6$ to $100$. Detailed descriptions of the test problems are provided in Appendix~\ref{app:cases}.

\subsection{Protocol}

\noindent For the single-fidelity SCBO baseline, the number of initial design points is chosen to match the total cost of initial points used in the multi-fidelity setting. For both MF-SCBO and SCBO, Matérn kernels with parameter $\nu = 2.5$ and anisotropic length scales are employed. The length scales are constrained to the interval $[0.005,\,4]$, and the observation noise is bounded in $[10^{-6},\,10^{-3}]$. As explained in \cite{eriksson_scalable_2021}, a bilog transformation ($\operatorname{bilog}(y) = \operatorname{sgn}(y)\ln(1 + |y|)$)
is applied to the constraints values in order to magnify the region around zero and thus emphasize sign changes that are decisive for feasibility. The total number of initial points for multi-fidelity methods are fixed to $4d$ with $d$ the dimension of the problem. In the multi-fidelity experiments, the scaling parameters are constrained to $\rho^{(s)} \in [1/2,2]$, rather than $[0,2]$. This choice is made to ensure that the low-fidelity information is effectively used by MF-SCBO, thereby enabling a comparison with the single-fidelity SCBO experiments based only on high-fidelity evaluations. For MF-MES, the original method was adapted to handle constrained multi-fidelity problems by introducing penalization terms for the constraints, with penalty parameter fixed to $100$. Nevertheless, the scaling to high dimensional problems was computationally difficult and we enforced a time limit of 120h on each run.\\

\noindent To compare feasible and infeasible solutions, we adopt the strategy of \cite{hernandez2016, eriksson_scalable_2021}, in which any feasible solution is considered preferable to an infeasible one. Thus, infeasible solutions are assigned to a default value equal to the largest observed objective value. In the plots, the mean is reported together with the $10^{\text{th}}$ and $90^{\text{th}}$ percentiles over $50$ independent runs, except for the \texttt{solar} problems \cite{andres2025} , for which only $5$ runs are performed due to their computational cost. In the \textit{predicted} cases, the plot is associated to the evaluation of the real objective $f(\cdot)$ and not the prediction $\mu^{S}(\cdot)$ used in the method in order to observe the real behavior. Due to the computational limitations of MF-MES, mainly arising from the optimization of the acquisition function in high-dimensional settings, experiments with this method are restricted to problems of dimension at most $8$, using $5$ independent runs.\\

\noindent A study of the impact of various method parameters, including the number of fidelity levels, batch size, and cost selection, is provided in Appendix~\ref{app:impact_params} for the synthetic test functions.

\subsection{Results}
\begin{figure}[htpb]
    \centering
    \includegraphics[width=1\linewidth]{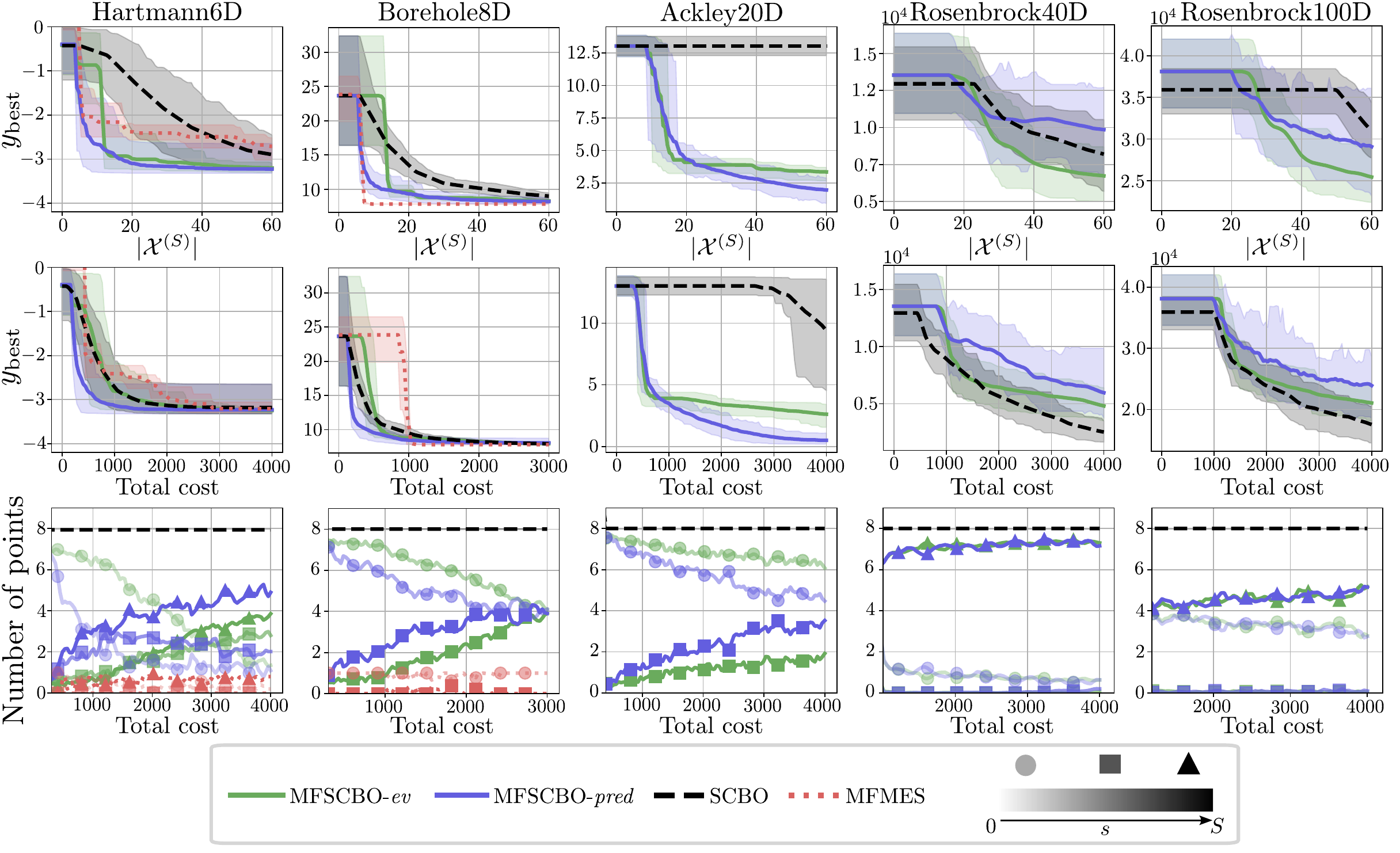}
     \caption{Convergence curves for the synthetic experiments: Hartmann6D, Borehole8D, Ackley20D, Rosenbrock40D, and Rosenbrock100D. Blue corresponds to the MF-SCBO-\textit{predicted} method, green to MF-SCBO-\textit{evaluated}, black to SCBO, and red to MFMFES. The first row shows the best value as a function of high-fidelity evaluations, the second row shows the best value versus total cost, and the third row displays the number of points used at each fidelity level $s$ for each iteration over the total cost.}
    \label{fig:synthetic_plots}
\end{figure}

\begin{figure}[htpb]
    \centering
    \includegraphics[width=0.8\linewidth]{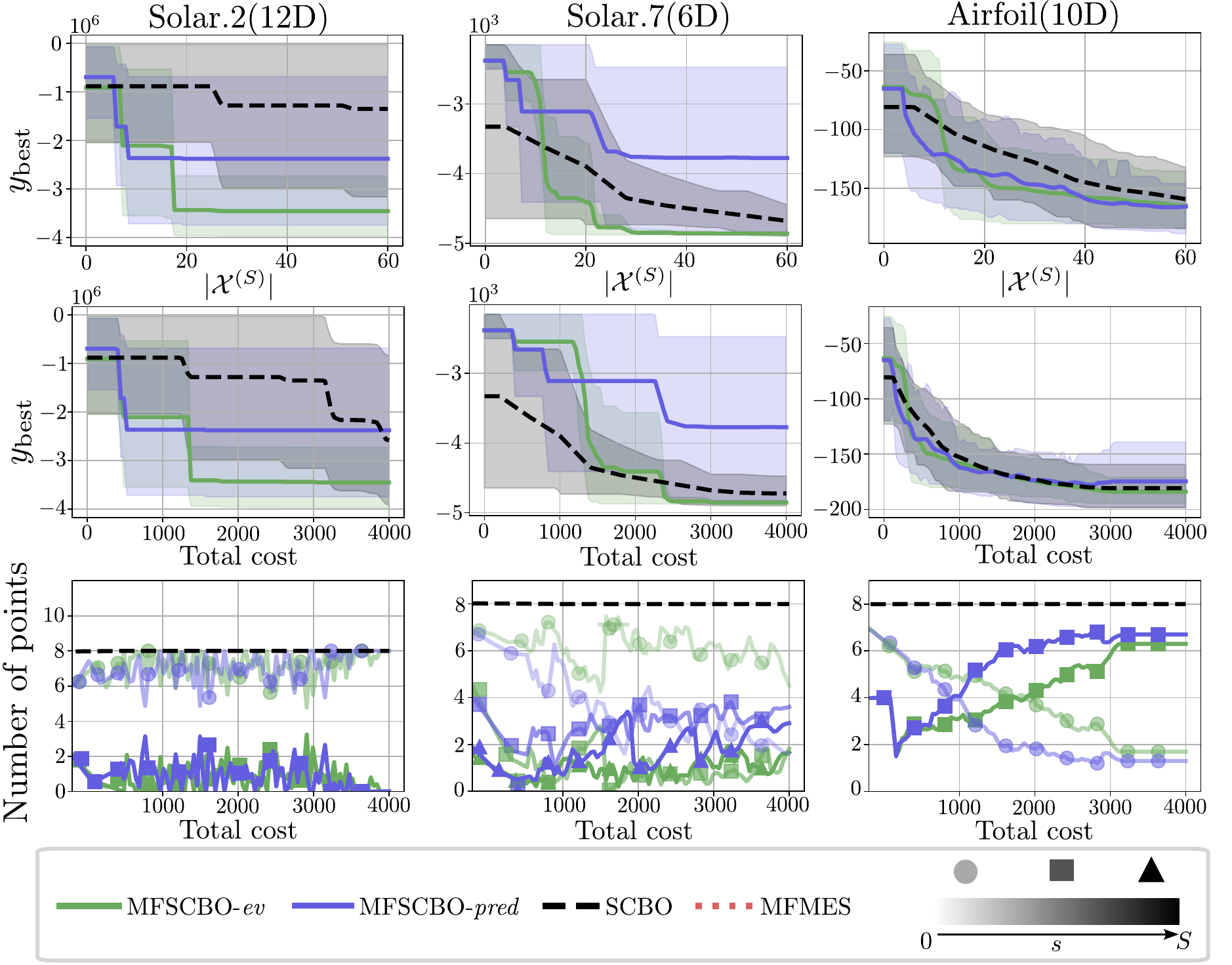}
     \caption{Convergence curves for the realistic experiments: Solar2, Solar7 and Airfoil. Blue corresponds to the MF-SCBO-\textit{predicted} method, green to MF-SCBO-\textit{evaluated}, black to SCBO, and red to MFMFES. The first row shows the best value as a function of high-fidelity evaluations, the second row shows the best value versus total cost, and the third row displays the number of points used at each fidelity level $s$ for each iteration over the total cost.}
    \label{fig:real_plots}
\end{figure}

In \Cref{fig:synthetic_plots} and \Cref{fig:real_plots}, convergence curves are reported both as a function of the number of high-fidelity evaluations and as a function of the total cost.  The number of samples drawn at  each fidelity level per iteration is also shown as a function of cost.  In \Cref{table:synthetic_table} and \Cref{table:real_table}, the mean of the optimal values and the percentiles over repeated runs are listed, together with the main experimental parameters, including batch size, fidelity-dependent costs, and the number of initial points.\\

\noindent In \Cref{fig:synthetic_plots}, results on synthetic test functions are presented. Overall, multi-fidelity methods outperform the single-fidelity SCBO baseline. For \texttt{Rosenbrock40D} and \texttt{Rosenbrock100D}, convergence appears weaker when measured against cost, while it remains more favorable when measured against the number of high-fidelity evaluations. This behavior is explained by the small ratio between fidelity costs: excessive use of low-fidelity evaluations can then outweigh the benefit of reducing high-fidelity evaluations in terms of total budget. A similar effect was observed in Appendix~\ref{app:impact_params}, where the influence of cost ratios on convergence was analyzed. In these particular cases, there is no tendency of a "phase transition" between low- and high-fidelity sampling over iterations, unlike the other cases. Furthermore, the medium-fidelity level seems to be almost never used here. Finally, the results differ from the other benchmarks, where the \textit{predicted} variant consistently outperforms the \textit{evaluated} variant.\\

\begin{table}[htpb]
\small
    \centering
    \setlength{\tabcolsep}{0.05pt} % default is ~6pt
    \begin{tabular}{c c | c c c c c}
    \hline
    \hline
     \multicolumn{2}{c|}{Instance}  & Hart.6D & Borehole8D & Ackley20D & Rosenb.40D & Rosenb.100D \\
     \hline
    $\begin{array}{c}
          \text{Parameters}
    \end{array}$ & $\begin{array}{c}
         \text{dim}  \\
         S+1\\
         \eta \\
         \text{batch size}\\
         \#\mathcal{X}_0
    \end{array}$
         & $\begin{array}{c}
         6 \\
         3\\
         (1,10,20) \\
         8\\
         (12,8,4)
    \end{array}$ & $\begin{array}{c}
         8 \\
         2\\
         (1,20) \\
         8\\
         (26, 6)
    \end{array}$ & $\begin{array}{c}
         20 \\
         2\\
         (1,20) \\
         8\\
         (72,8)
    \end{array}$ & $\begin{array}{c}
         40 \\
         3\\
         (1,10,20) \\
         8\\
         (104,40,16)
    \end{array}$ & $\begin{array}{c}
         100 \\
         3\\
         (1,10,20) \\
         8\\
         (200,40,20)
    \end{array}$
      \\
      \hline 
      $\begin{array}{c}
          \text{Results}
    \end{array}$ & $\begin{array}{c}
         \text{MFSCBO-\textit{eval.}}  \\
         \text{MFSCBO-\textit{pred.}}\\
         \text{SCBO} \\
         \text{MFMEF}
    \end{array}$
         & $\begin{array}{c}
         \mathbf{-3.25}(\pm0.05) \\
         -3.23(\pm0.06)\\
         -3.18(\pm0.13) \\
         -3.20(\pm 0.11)
    \end{array}$ &  $\begin{array}{c}
         7.94(\pm0.03) \\
         \mathbf{8.07}(\pm0.11)\\
         7.95(\pm0.03) \\
         7.83(\pm0.01)
    \end{array}$ & $\begin{array}{c}
         2.62(\pm0.20) \\
         \mathbf{0.50}(\pm0.21)\\
         9.40(\pm1.07) \\
         -
    \end{array}$ & $\begin{array}{c}
         4.81(\pm0.32)\!\times\!10^3 \\
        5.97(\pm0.62)\!\times\! 10^3\\
         \mathbf{2.52}(\pm0.21)\!\times\! 10^3 \\
         -
    \end{array}$ & $\begin{array}{c}
         2.11(\pm0.06)\!\times\! 10^4 \\
        2.39(\pm0.14)\!\times\! 10^4\\
         \mathbf{1.75}(\pm0.07)\!\times\!10^4 \\
         -
    \end{array}$
      \\
          \hline 
          \hline 
    \end{tabular}
\caption{Parameters and optimization results for the synthetic functions: Hartmann6D, Borehole8D, Ackley20D, Rosenbrock40D, and Rosenbrock100. The table lists the problem dimension, number of fidelities, associated cost, batch size, and number of initial points for each fidelity. The mean of the best value after $4000$ objective function evaluations is reported for each method, along with the $1.96\times$ standard deviation. The best value across methods is highlighted in bold.}
    \label{table:synthetic_table}
\end{table}

\noindent  In \Cref{fig:real_plots}, we report the results on the more realistic benchmark problems. As expected, the \texttt{solar} problems are particularly challenging because they include failure points, referred to as \textit{hidden constraints} in \cite{LeDigabel2023,andres2025}. These points emulate simulation failures or numerical errors that may occur during the optimization process and are handled by assigning a large penalty value to either the objective function or the violated constraints. The difficulty is also increased by the large number of constraints, ranging from $6$ to $12$. Consequently, MFMES fails to converge on the \texttt{solar7} problem. In contrast, the proposed multi-fidelity approach achieves good performance on \texttt{solar2}. For \texttt{solar7}, the \textit{evaluated} variant exhibits slightly faster convergence during the first $20$ high-fidelity evaluations, whereas the \textit{predicted} variant performs poorly. This behavior is likely due to less accurate high-fidelity predictions caused by a larger number of infeasible points in \texttt{solar7} \cite{andres2025}. For the airfoil shape optimization problem, the multi-fidelity approach converges more rapidly than its single-fidelity counterpart, measured as a function of the number of high-fidelity evaluations. The optimization process also exhibits a transition in the fidelity levels used throughout the search. Early iterations rely mostly on low-fidelity evaluations to explore the design space, while later iterations increasingly high-fidelity evaluations to refine the solution.

\begin{table}[htpb]
\small
    \centering
    \setlength{\tabcolsep}{0.05pt} % default is ~6pt
    \begin{tabular}{c c | c c c c}
    \hline
    \hline
     \multicolumn{2}{c|}{Instance}  & Solar2 & Solar7 & Airfoil  \\
     \hline
    $\begin{array}{c}
          \text{Parameters}
    \end{array}$ & $\begin{array}{c}
         \text{dim}  \\
         S+1\\
         \eta \\
         \text{batch size}\\
         \#\mathcal{X}_0
    \end{array}$
         & $\begin{array}{c}
         12 \\
         3\\
         (1,50) \\
         8\\
         (42, 6)
    \end{array}$ & $\begin{array}{c}
         6 \\
         3\\
         (1,50) \\
         8\\
         (20, 4)
    \end{array}$ & $\begin{array}{c}
         10 \\
         2\\
         (1,20) \\
         8\\
         (36,4)
    \end{array}$
      \\
      \hline 
      $\begin{array}{c}
          \text{Results}
    \end{array}$ & $\begin{array}{c}
         \text{MFSCBO-\textit{eval.}}  \\
         \text{MFSCBO-\textit{pred.}}\\
         \text{SCBO} \\
         \text{MFMEF}
    \end{array}$
         & $\begin{array}{c}
         \mathbf{-3.46}(\pm0.82)\!\times\!10^6 \\
         -2.34(\pm1.89)\!\times\!10^6\\
         -2.59(\pm1.97)\!\times\!10^6 \\
         -
    \end{array}$ &  $\begin{array}{c}
         \mathbf{-4.86}(\pm0.023)\!\times\! 10^3 \\
         -3.77(\pm0.944)\!\times\!10^3\\
         -4.80(\pm0.178)\!\times\! 10^3 \\
         2.77(\pm1.98)\!\times\! 10^9 
    \end{array}$ & $\begin{array}{c}
         \mathbf{-1.84}(\pm0.03)\!\times\! 10^2 \\
         -1.75(\pm0.06)\!\times\! 10^2\\
         -1.81(\pm0.05)\!\times\!10^2 \\
         -
    \end{array}$\\
    \hline
    \hline
    \end{tabular}
\caption{Parameters and optimization results for the realistic functions: Solar2, Solar7 and Airfoil. The table lists the problem dimension, number of fidelities, associated cost, batch size, and number of initial points for each fidelity. The mean of the best value after $4000$ objective function evaluations is reported for each method, along with the $1.96\times$ standard deviation. The best value across methods is highlighted in bold.}
    \label{table:real_table}
\end{table}

\section{Conclusion}\label{sec:conclusion}

We considered multi-fidelity Bayesian optimization in a high-dimensional setting, with an arbitrary number of fidelity levels and black-box constraints. To address this problem, we proposed \textit{Multi-Fidelity Scalable Constrained Bayesian Optimization} (MF-SCBO), a direct extension of the SCBO framework \cite{eriksson_scalable_2021}. Two variants were introduced. The first updates the trust-region center using only high-fidelity evaluations, while the second uses predictions of the highest-fidelity model based on data from all fidelity levels. To improve the robustness of the latter approach when these predictions are inaccurate, a leave-one-out criterion is used to maintain consistency in the choice of the trust-region center. This criterion can be computed analytically and thus introduces only a negligible additional computational cost.\\

MF-SCBO was evaluated on a range of problems, from standard synthetic benchmarks to realistic engineering applications. Overall, it achieved better convergence than both the single-fidelity SCBO algorithm and MF-MES. In particular, MF-MES quickly became computationally impractical as the problem dimension increased, whereas MF-SCBO remained applicable to high-dimensional problems. When the high-fidelity predictions were sufficiently accurate, the \textit{predicted} variant generally provided faster convergence than the \textit{evaluated} variant, as observed for the Hartmann, Borehole, and Ackley problems. This was not the case for the Rosenbrock and Solar problems, where the predictions were less reliable. Finally, the influence of several optimization parameters, including the batch size, the number of fidelity levels, and the relative evaluation costs, was investigated.\\

Future work could focus on extending MF-SCBO to more general multi-source Bayesian optimization settings \cite{Poloczek2017}, where different sources may have different costs, accuracies, and biases, without assuming that the error decreases and the computational cost increases monotonically with the fidelity level. In such settings, the most appropriate source may also depend on the location in the input space, as different sources may provide more accurate or informative predictions in different regions. Another challenging direction would be to consider constraints that depend on the fidelity level, which would require extending the proposed framework to account for different feasible regions across sources. 
The current approach could be extended to take into account the remaining evaluation budget, for example through a non-myopic strategy that considers the potential value of future evaluations as in \cite{difiore2024}. Finally, scalability to even higher dimensional problems could be improved by considering alternative GP models, such as additive Gaussian processes, which decompose the objective into lower-dimensional components.\\

{\centering \bf Acknowledgement}
The authors acknowledge the support of the French Agence Nationale de la Recherche (ANR), under Grant No. ANR-21-CE45-0013, Project NEMO. 

\appendix

\section{Cases}\label{app:cases}
\paragraph{Hartmann6D.}
The Hartmann function family is commonly used to benchmark optimization algorithms. Here, the six-dimensional case is considered, which has six local minima. The global minimum is located at $\boldsymbol{x}^{*}=\begin{bmatrix}0.20169 &0150011 &0.476874 &0.275332 &0.311652&0.6573\end{bmatrix}^\top$ where the function value is $-3.32237$. The computational domain is defined as $\Omega=[0,1]^6$. The multi-fidelity formulation studied in \cite{kandasamy2019} defines the $s$-th fidelity level as
\begin{equation*}
    f^{(s)}(\boldsymbol{x}) = \sum_{i=1}^4 \alpha_i^{(s)} 
    \exp\left(-\sum_{j=1}^6 A_{ij}(x_j - P_{ij})^2\right),
\end{equation*}
where $A, P \in \mathbb{R}^{4 \times 6}$ are fixed matrices defined as
\begin{equation*}
    A :=
    \begin{bmatrix}
    10 & 3 & 17 & 3.5 & 1.7 & 8 \\
    0.05 & 10 & 17 & 0.1 & 8 & 14 \\
    3 & 3.5 & 1.7 & 10 & 17 & 8 \\
    17 & 8 & 0.05 & 10 & 0.1 & 14
    \end{bmatrix}, 
    \qquad
    P := 10^{-4}
    \begin{bmatrix}
    1312 & 1696 & 5569 & 124 & 8283 & 5886 \\
    2329 & 4135 & 8307 & 3736 & 1004 & 9991 \\
    2348 & 1451 & 3522 & 2883 & 3074 & 6650 \\
    4047 & 8828 & 8732 & 5743 & 1091 & 381
    \end{bmatrix}.
\end{equation*}
The fidelity-dependent parameter vector $\boldsymbol{\alpha}^{(s)}$ is defined as
\begin{equation*}
    \boldsymbol{\alpha}^{(s)} := \boldsymbol{\alpha} + (S - s)\boldsymbol{\delta},
    \qquad \text{with} \qquad
    \begin{cases}
    \boldsymbol{\alpha} := 
    \begin{bmatrix}
    1 & 1.2 & 3 & 3.2
    \end{bmatrix}^\top, \\[0.3em]
    \boldsymbol{\delta} :=
    \begin{bmatrix}
    0.01 & -0.01 & -0.1 & 0.1
    \end{bmatrix}^\top.
    \end{cases}
\end{equation*}
Following \cite{letham2019}, a single constraint is imposed to restrict feasible points to lie inside the unit sphere: $\|\boldsymbol{x}\| \leq 1$.

\paragraph{Borehole8D.}
The Borehole function is a height-dimensional function, often used in Bayesian optimization, which models water flow through a borehole. The computational domain is defined as 
\begin{multline*}
\Omega = [0.05,0.15] \times [100,5\times 10^4] \times [6.307\times 10^4, 1.156\times 10^{5}] \times [990,1110] \\\times [63.1,116] \times [700,820] \times [1120,1680] \times [9855,12045].
\end{multline*}
As used in \cite{kandasamy2019}, two fidelity levels are considered and defined by 
\begin{align*}
f^{(1)}(\boldsymbol{x}) &= 
\frac{2\pi\, x_3 (x_4 - x_6)}
{\ln(x_2/x_1)\left(1 + \frac{2 x_7 x_3}{\ln(x_2/x_1)\, x_1^2 x_8} + \frac{x_3}{x_8}\right)},\\
f^{(0)}(\boldsymbol{x}) &= 
\frac{5\, x_3 (x_4 - x_6)}
{\ln(x_2/x_1)\left(1.5 + \frac{2 x_7 x_3}{\ln(x_2/x_1)\, x_1^2 x_8} + \frac{x_3}{x_8}\right)}, \\
\end{align*}
where $f^{(1)}$ is the classical Borehole function. No constraints are considered, following the standard benchmark setting in the literature.

\begin{figure}[htpb]
    \centering
    \includegraphics[width=0.8\linewidth]{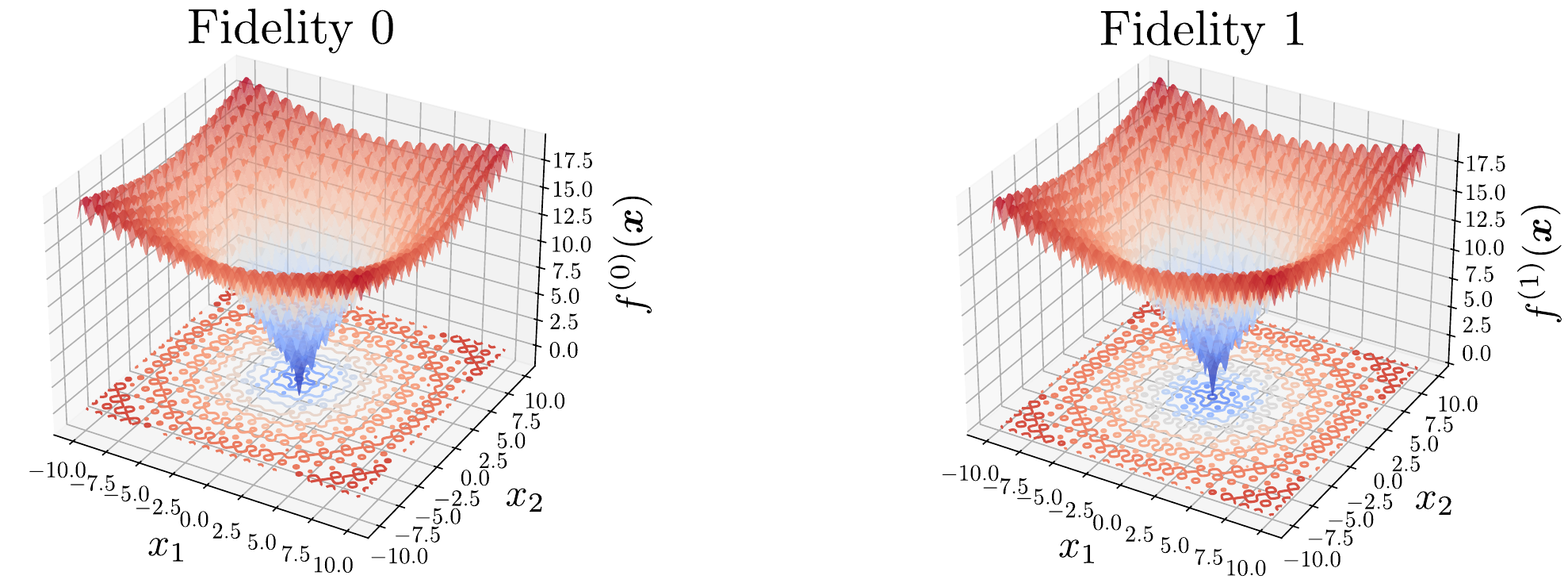}
    \caption{Two dimensional Ackley function at two different fidelity levels, $0$ for the low fidelity function and $1$ for the  high fidelity function.}
    \label{fig:ackley_illu}
\end{figure}

\paragraph{Ackley20D.}
The Ackley function is a classical function, proposed by David Ackley, to test the performance of optimization algorithms. Its global minimum is located at the origin where the function value is zero. However, the presence of a huge number of local minima makes the problem very difficult. The computational domain considered is defined as $\Omega =[-5,10]^{d}$ with $d=20$. Two fidelity levels are considered, following  \cite{agrawal2025}, and given by
\begin{align*}
f^{(1)}(\boldsymbol{x}) &= -20 \exp\left(-0.2 \sqrt{\frac{1}{d}} \|\boldsymbol{x}\|_2 \right)
- \exp\left(\frac{1}{d} \sum_{i=1}^{d} \cos(2\pi x_i)\right)
+ 20 + \exp(1), \\
f^{(0)}(\boldsymbol{x}) &= -22 \exp\left(-0.2 \sqrt{\frac{1.1}{d}} \|\boldsymbol{x}\|_2 \right)
- 0.9 \exp\left(\frac{1}{d} \sum_{i=1}^{d} \cos(2\pi x_i)\right)
+ 20 + \exp(1),\\
\end{align*}
where $f^{(1)}$ is the classical Ackley function. In \Cref{fig:ackley_illu}, the two-dimensional case is illustrated for the two fidelity levels, where the global minimum in the low-fidelity model is displaced from the origin. Two constraints are imposed, following \cite{eriksson_scalable_2021}: 
\begin{equation*}
c_1(\boldsymbol{x}) = \sum_{i=1}^{20} x_i \leq 0,
\qquad
c_2(\boldsymbol{x}) = \|\boldsymbol{x}\|_2 - 5 \leq 0,
\end{equation*}
the first is a hyperplane constraint, and the second defines a sphere centered at the origin with radius $5$.

\begin{figure}[htpb]
    \centering
    \includegraphics[width=1\linewidth]{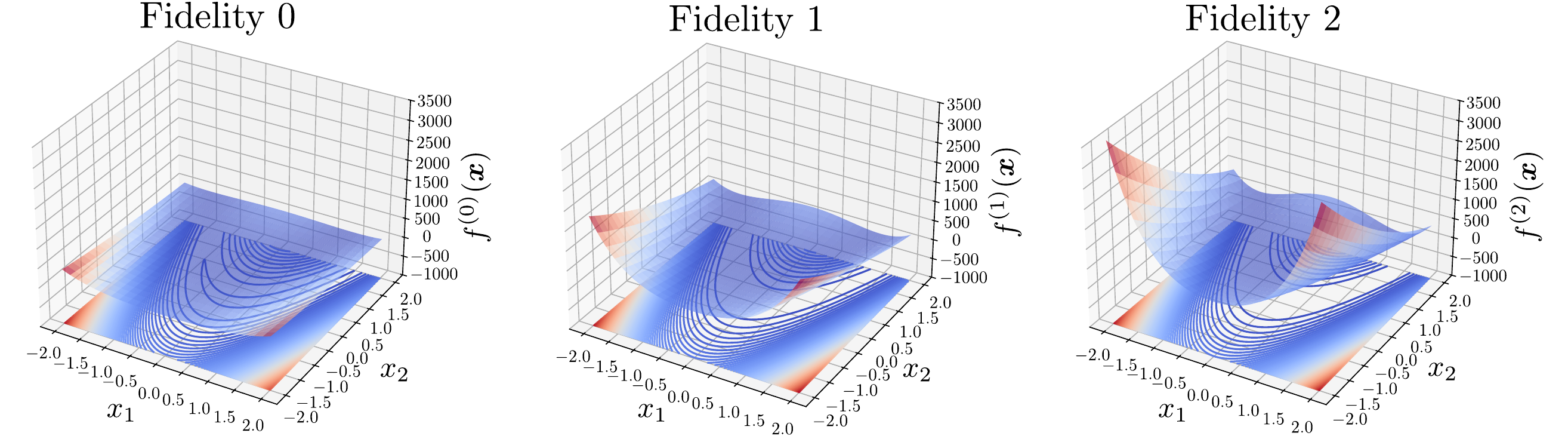}
    \caption{Two dimensional Rosenbrok function at three different fidelity levels, from  $0$ (lowest) to $2$ (highest).}
    \label{fig:rosenbrock_illu}
\end{figure}

\paragraph{Rosenbrock40D and Rosenbrock100D.} The Rosenbrock function, also known as the Banana function, is a classical benchmark in gradient-based optimization. Its global minimum is located at $\boldsymbol{x}^*=\begin{bmatrix}
    1 & \ldots & 1
\end{bmatrix}^\top$ where the function value is zero. The minimum lies inside a parabolic shaped flat valley, which makes convergence difficult, particularly in high-dimensional settings. Note that there is also local minimums. The computational domain is defined as $\Omega=[-2,2]^d$ with $d\in\{40,100\}$. Three fidelity levels are considered, following \cite{mainini2025}, and defined by 
\begin{align*}
f^{(2)}(\boldsymbol{x})&=\sum_{i=1}^{d-1}100(x_{i+1}-x_i^2)^2+(1-x_i)^2,\\
f^{(1)}(\boldsymbol{x})&=\sum_{i=1}^{d-1}50(x_{i+1}-x_i^2)^2+(-2-x_i)^2-0.5\sum_{i=1}^d x_i,\\ 
f^{(0)}(\boldsymbol{x})&=\frac{f^{(2)}(\boldsymbol{x})-4-0.5\sum_{i=1}^dx_i}{10+0.25\sum_{i=1}^dx_i},
\end{align*}
where $f^{(2)}$ is the classical Rosenbrock function. \Cref{fig:rosenbrock_illu} illustrates the three fidelities level in the two-dimensional case. The constraints considered are :
\begin{equation*}
    c_1(\boldsymbol{x})=-\sum_{i=1}^d \cos(x_i)+1\leq 0, \qquad c_2(\boldsymbol{x})=\sum_{i=1}^dx_i\leq0.
\end{equation*}

\paragraph{Solar2 and Solar7.}
\texttt{solar} is a solar thermal power plant simulator designed for blackbox optimization benchmarking \cite{andres2025}. Several test cases are available; here, \texttt{solar2} and \texttt{solar7} are considered.\\

\noindent In \texttt{solar2}, the objective is to minimize the heliostat field surface using $14$ variables, of which $2$ are discrete, subject to $12$ constraints. In \texttt{solar7}, the objective is to maximize the receiver efficiency over a $24$-hour period. This problem involves $7$ variables, including one discrete variable, and $6$ constraints. Since the discrete variables are not bounded over $\mathbb{N}$, they are fixed to the initial values provided by \texttt{solar}. The characteristics are summarized in \Cref{table:solar_charact}.\\

\begin{table}[htpb]
    \centering
    \begin{tabular}{c|c c c|c c|c}
    \hline
    \hline
     Instance & \multicolumn{3}{c|}{\# of variables} 
         & \multicolumn{2}{c|}{\# of constraints} 
         & \# of stoch. outpus \\
         %\cline{2-7}
          & cont. & discr. & fixed & simu. & a priori & (obj. or constr.) \\
         \hline
          
         \texttt{solar2} & $12$ & $2$ & $(x_6,\;x_{11})=(2650,\;36)$& $7$ & $5$ & $4$\\
          \texttt{solar7} & $6$ & $1$ & $x_4=40$& $4$ & $2$ & $3$\\  
          \hline 
          \hline 
    \end{tabular}
    \caption{Characteristics of the \texttt{solar2} and \texttt{solar7} instances, including the number of variables (continuous and discrete), the fixed values of the discrete variables, the number of constraints (simulated and a priori), and the number of stochastic outputs.}
    \label{table:solar_charact}
\end{table}

\noindent Hidden constraints are introduced to represent failures encountered in real blackbox settings. In such cases, the simulation return a value of $10^{20}$. In this case, some simulation output return the value $10^{20}$. Not all outputs are necessarily violated. Violations occur less frequently at higher fidelity levels (closer to level $1$). Moreover, some outputs rely on surrogate approximations and are therefore stochastic. To ensure deterministic evaluations, the random seed is fixed to $0$ for both problems.

\begin{figure}[htpb]
    \centering
    \includegraphics[width=0.6\linewidth]{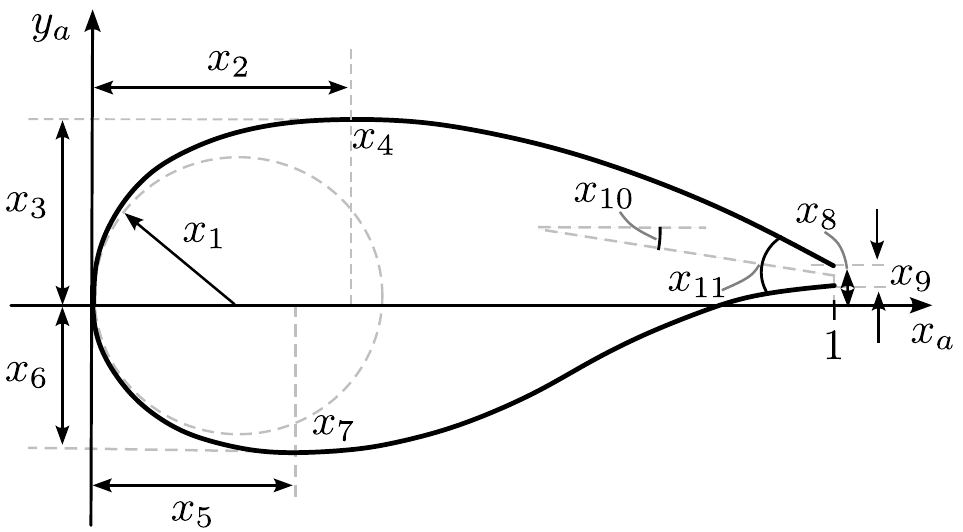}
\caption{Illustration of the PARSEC airfoil parameterization.}
    \label{fig:parsec_airfoil}
\end{figure}

\paragraph{Airfoil shapes.}
Airfoil shape optimization is a a standard and complex optimization problem in aerodynamic design. Here, the maximization of the lift-to-drag ratio using the PARSEC parameterization is considered \cite{sobieczky1999,do2025}. \Cref{fig:parsec_airfoil} illustrates the PARSEC geometry, and the associated parameters \cite{shinkyu2005} are listed in \Cref{table:parsec_param}. As in \cite{do2025}, the trailing-edge thickness is fixed to $x_9=0$.  

\begin{table}[htpb]
\label{table:parsec_param} 
\begin{center}
  \begin{tabular}{c c l}
  \hline
\hline
     Coefficient &   Bound  & Description \\ 
     \hline
     $x_1$  &  $[0.005, 0.06]$ &Leading-edge radius \\
     $x_2$ & $[0.25, 0.5]$&Upper crest position in horizontal coordinates \\
     $x_3$ & $[0.05, 0.15]$&Upper crest position in vertical coordinates \\
     $x_4$ & $[-1, -0.4]$&Upper crest curvature \\
     $x_5$ & $[0.35, 0.5]$&Lower crest position in horizontal coordinates \\
     $x_6$ & $[-0.12, -0.04]$&Lower crest position in vertical coordinates \\
     $x_7$ & $[0.3, 1]$&Lower crest curvature \\
     $x_8$ & $[-0.02, 0.02]$&Trailing-edge vertical coordinate \\
     $x_9$ & $[0, 0]$&Trailing-edge thickness \\
     $x_{10}$ & $[-8, -3]$&Trailing-edge direction \\
     $x_{11}$ & $[4, 8]$&Trailing-edge wedge angle \\
    \hline
\hline
  \end{tabular}
  \end{center}
  \caption{Description of the shape variables for the PARSEC parameterization of airfoils. The trailing-edge thickness, $x_9$, is not optimized and is fixed to $0$.}
\end{table}
\noindent Two fidelity levels are considered. The high-fidelity function relies on a multi-point design strategy to ensure good aerodynamic performance across several operating conditions, whereas the low-fidelity function considers only a single condition \cite{do2025}. Four operating conditions (see \Cref{table:airfoil_param}) are selected to obtain shapes that are robust with respect to variations in speed and angle of attack at a fixed Reynolds number. 
\begin{table}[htpb]
\begin{center}
  \begin{tabular}{c c c c c}
  \hline
\hline
     Index &   Reynolds  & Mach & Angle of attack (deg.) & Weights $\omega_i$ \\ 
     \hline
     $1$  &  $6.3\times 10^6$ & $0.5$ & $2$ & $0.4$\\
     $2$ & $6.3\times 10^6$& $0.55$ & $2.2$ & $0.2$\\
     $3$ & $6.3\times 10^6$& $0.5$  & $2.5$ & $0.2$\\
     $4$ & $6.3\times 10^6$& $0.55$ & $2$ & $0.2$\\
    \hline
\hline
  \end{tabular}
  \end{center}
  \caption{Operating conditions used in the multi-point design strategy, together with the associated weights $\omega_i$.}
  \label{table:airfoil_param} 
\end{table}
\noindent The objective functions are defined as
\begin{equation*}
    f^{(1)}(\boldsymbol{x}) = -\sum_{i=1}^4 \omega_i \frac{c_{\text{L},i}}{c_{\text{D},i}}, 
    \qquad 
    f^{(0)}(\boldsymbol{x}) = -\frac{c_{\text{L},1}}{c_{\text{D},1}},
\end{equation*}
where $c_{\text{L},i}$ and $c_{\text{D},i}$ denote the lift and drag coefficients under the $i$-th condition, and $\omega_i$ are the weights given in \Cref{table:airfoil_param}. The constraint considered is a bound on the moment coefficient of the first condition, $c_{\text{M},1}$, which applies to both high- and low-fidelity levels, defined as :
\begin{equation*}
c_1(\boldsymbol{x})=c_{\text{M},1}+0.05\leq 0, \qquad c_2(\boldsymbol{x})=-c_{\text{M},1}-0.1\leq 0.
\end{equation*}
Since certain shapes can cause the solver to fail, a default value of $10^2$ is assigned to both the objective and the constraints in such cases.\\

\noindent The upper and lower surface coordinates, $y_{\text{up}}$ and $y_{\text{low}}$, are described by PARSEC polynomial expansions \cite{dellavecchia2014}, depending of the shape variable listed in \Cref{table:airfoil_param} :
\begin{equation*}
    y_{\text{up}}(x_a) = \sum_{i=1}^6 a_{\text{up},i} x_a^{i-1/2}, 
    \qquad 
    y_{\text{low}}(x_a) = \sum_{i=1}^6 a_{\text{low},i} x_a^{i-1/2},
\end{equation*}
with $x_a \in [0,1]$. The coefficient vectors $\boldsymbol{a}_{\text{up}} \in \mathbb{R}^6$ and $\boldsymbol{a}_{\text{low}} \in \mathbb{R}^6$ are obtained by solving
\begin{equation*}
C(x_2)\boldsymbol{a}_{\text{up}}=\boldsymbol{b}(x_1,x_3,x_4,x_8,x_9,x_{10},-x_{11}), 
\qquad 
C(x_5)\boldsymbol{a}_{\text{low}}=\boldsymbol{b}(x_1,x_6,x_7,x_8,-x_9,x_{10},x_{11}),
\end{equation*}
where
\begin{equation*}
    C(z) = \begin{bmatrix}
        1 & 1 & 1 & 1 & 1 & 1\\
        z^{1/2} & z^{3/2} & z^{5/2} & z^{7/2} & z^{9/2} & z^{11/2}\\
        1/2 & 3/2 & 5/2 & 7/2 &9/2 &11/2\\
        \frac{1}{2}z^{-1/2} & \frac{3}{2}z^{1/2}& \frac{5}{2}z^{3/2} & \frac{7}{2}z^{5/2} & \frac{9}{2}z^{7/2} & \frac{11}{2}z^{9/2} \\
         -\frac{1}{4}z^{-3/2} & \frac{3}{4}z^{-1/2}& \frac{15}{4}z^{1/2} & \frac{35}{4}z^{3/2} & \frac{63}{4}z^{5/2} & \frac{99}{4}z^{7/2}\\
         1 & 0 & 0 & 0 & 0 & 0 
    \end{bmatrix}, 
    \qquad 
    \boldsymbol{b}(\boldsymbol{z})=
    \begin{bmatrix}
    z_4 + z_5/2\\
    z_2\\
    \tan(z_6 + z_7/2)\\
    0\\
    z_3\\
    \sqrt{2z_1}
    \end{bmatrix}.
\end{equation*}
\noindent To compute the lift and drag coefficients at each operating condition, the \texttt{XFOIL} solver \cite{mark1989} is employed.

\section{Impact of parameters: number of fidelity levels, batch size and cost}\label{app:impact_params}
In the various analyses conducted here, we will consider the parameters listed in \Cref{table:synthetic_table} unless otherwise specified.\\

\begin{figure}[htpb]
    \centering    \includegraphics[width=0.6\linewidth]{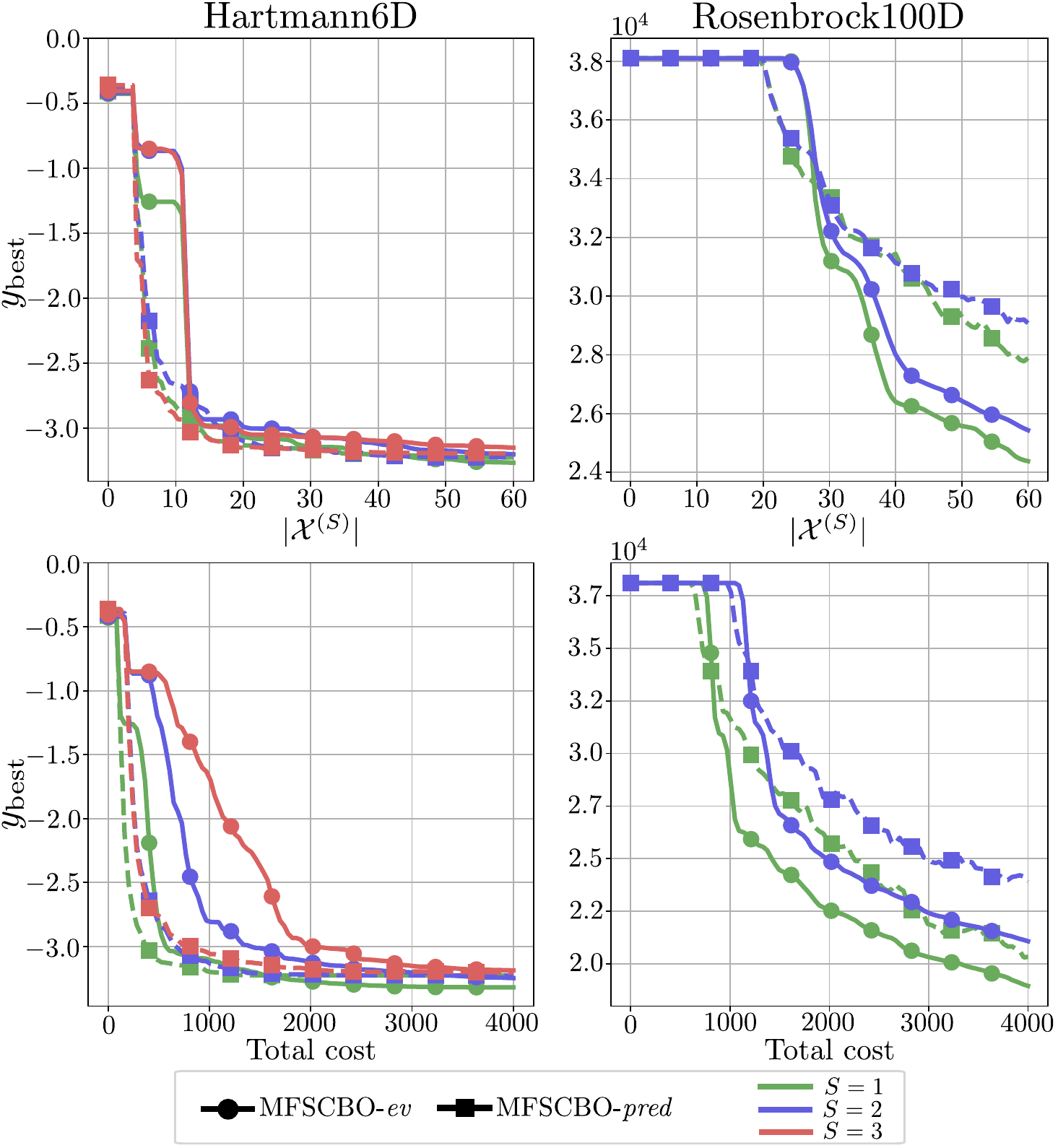}
     \caption{Convergence curves for different number of fidelity level for Hartmann6D and Rosenbrock100D. Green, blue and red correspond respectively to $2$, $3$ et $4$ number of fidelity levels. Circles correspond to the \textit{evaluated} method and squares to the \textit{predicted} method.}
    \label{fig:impact_fid}
\end{figure}

\noindent {\bf Number of fidelity levels :}
In this section, the impact of the number of fidelity levels is examined on the Hartmann6D and Rosenbrock100D functions. For the Hartmann6D case, three fidelity levels are examined: $2$, $3$, and $4$, which have initial point counts of $(20, 4)$, $(12, 8, 4)$, and $(10, 6, 4, 4)$, respectively. In the Rosenbrock100D case, two fidelity levels are studied, $2$ and $3$, which have initial point counts of $(240, 20)$ and $(200, 40, 20)$, respectively. The results are presented in \Cref{fig:impact_fid}. A high fidelity level appears to slow down convergence, particularly for the \textit{evaluated} mode and especially depending on the cost.

\begin{figure}[htpb]
    \centering    \includegraphics[width=1\linewidth]{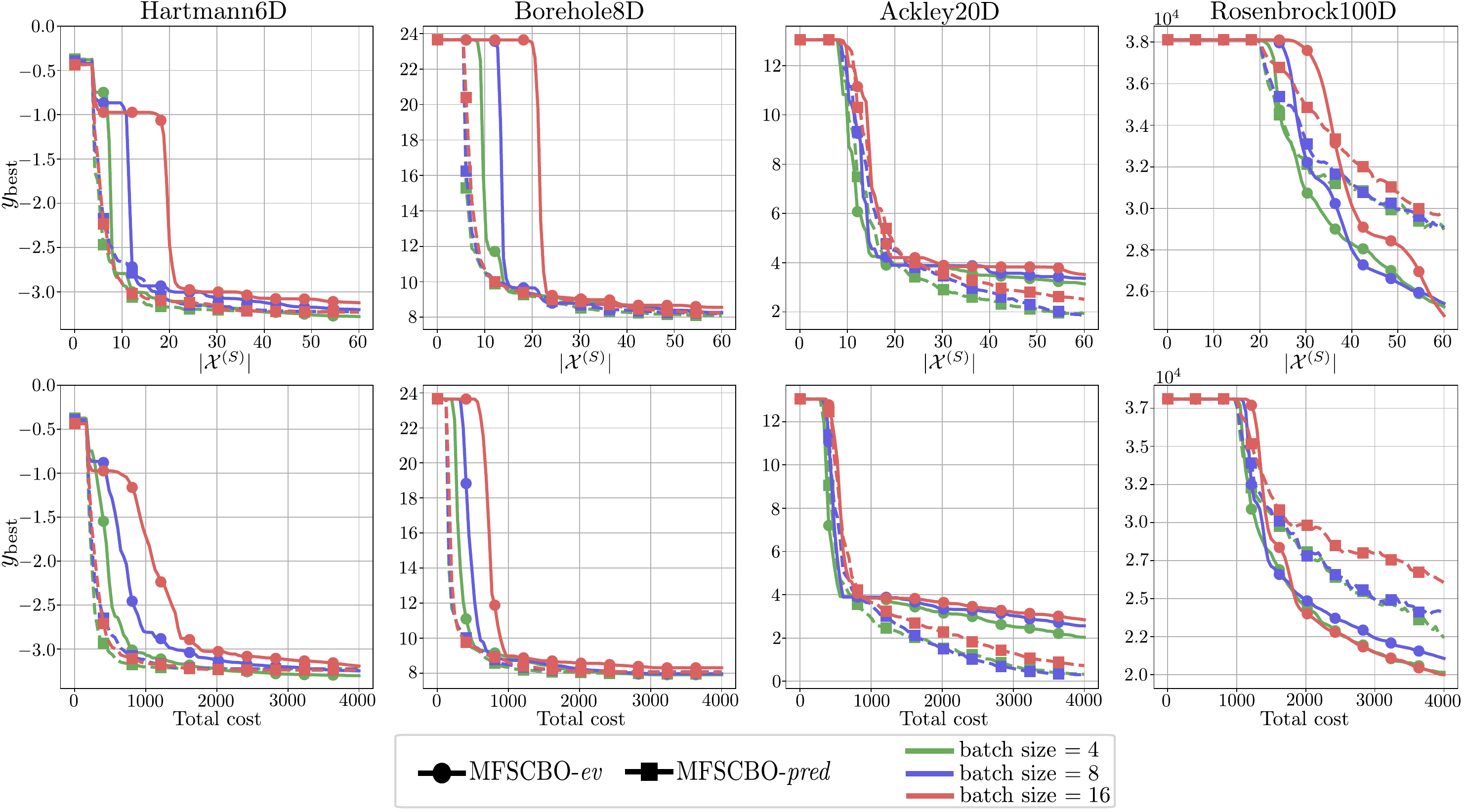}
     \caption{Convergence curves for different batch size for Hartmann6D, Borehole8D, Ackley20D and Rosenbrock100D. Green, blue and red correspond respectively to $4$, $8$ et $16$ batch sizes. Circles correspond to the \textit{evaluated} method and squares to the \textit{predicted} method.}
    \label{fig:impact_batch}
\end{figure}

\noindent {\bf Batch size :}
In this section, the impact of batch size is examined on the Hartmann6D, Borehole8D, Ackley20D, and Rosenbrock100D functions. We consider three different batch sizes: $4$, $8$, and $16$. The results are presented in \Cref{fig:impact_batch}. Batch size has a less significant effect in the \textit{predicted} case than in the \textit{evaluated} case. The observed discrepancy—which increases as the batch size increases—stems from the fact that, in the \textit{evaluated} case, the confidence region is updated only after the batch has been filled with points of the highest fidelity level.\\

\begin{figure}[htpb]
    \centering    \includegraphics[width=1\linewidth]{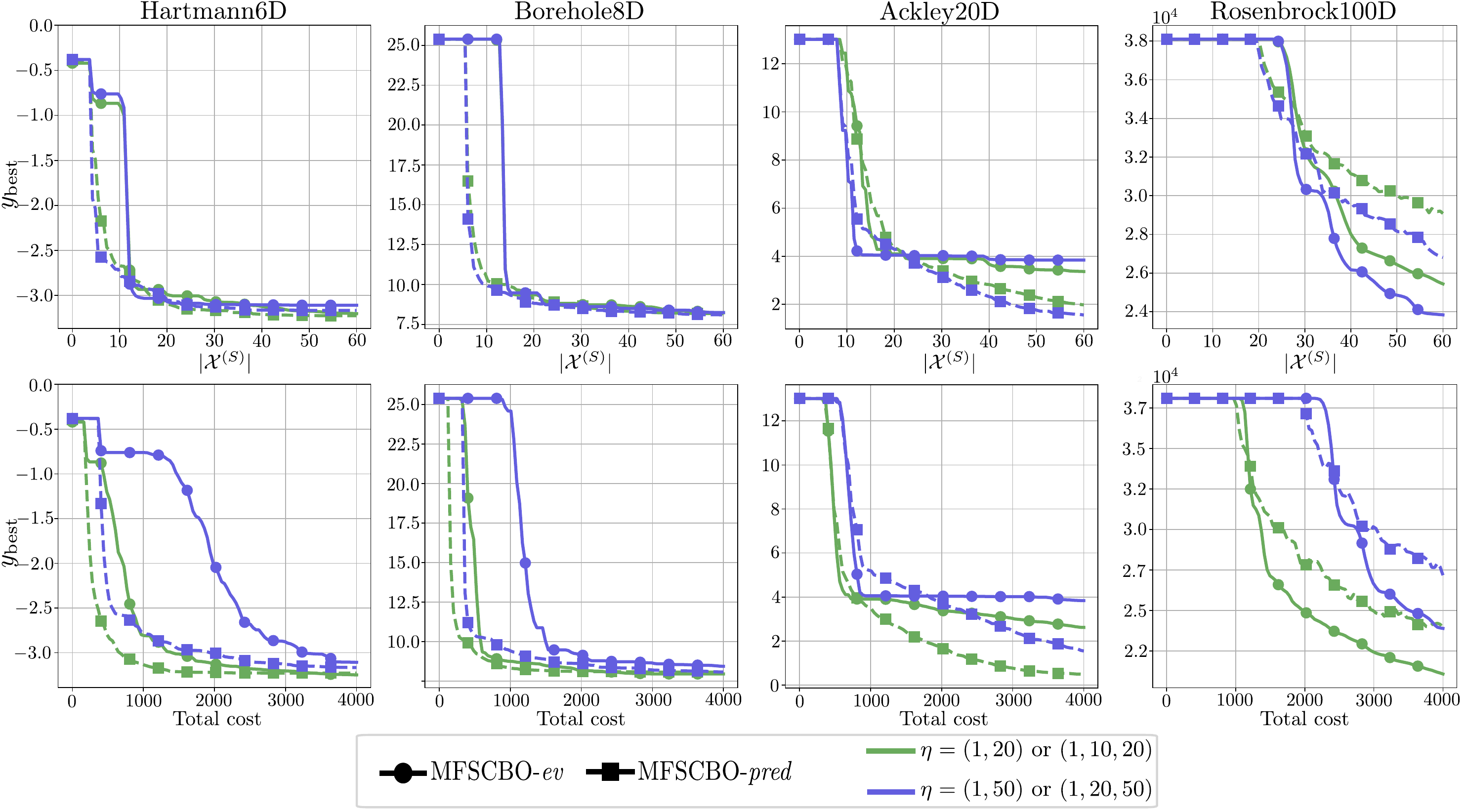}
     \caption{Convergence curves for different costs for Hartmann6D, Borehole8D, Ackley20D and Rosenbrock100D. Green and blue correspond respectively to $(1,20)$ and $(1,50)$ costs for two fidelities and $(1,10,20)$, $(1,20,50)$ for three fidelities. Circles correspond to the \textit{evaluated} method and squares to the \textit{predicted} method.}
    \label{fig:impact_costs}
\end{figure}

\noindent {\bf Cost :}
In this section, the impact of cost is examined for the Hartmann6D, Borehole8D, Ackley20D, and Rosenbrock100D functions. Two different cost levels are considered for each function: for the two fidelity levels $(1,20)$ and $(1,50)$; for three accuracy levels: $(1,10,20)$ and $(1,20,50)$. The results are presented in \Cref{fig:impact_costs}. The higher the cost, the greater the number of low-accuracy evaluations required to make it worthwhile to evaluate a point at a higher accuracy level. This is due to the criterion \eqref{eq:assignfid}, which weights the error made against the associated cost. It is this discrepancy that is observed in \Cref{fig:impact_costs} between the \textit{evaluated} and \textit{predicted} values.

\newpage

\bibliographystyle{abbrv}
\bibliography{biblio} 

@article{difiore2024,
title = {{NM2-BO}: Non-Myopic Multifidelity {Bayesian} Optimization},
journal = {Knowledge-Based Systems},
volume = {299},
pages = {111959},
year = {2024},
issn = {0950-7051},
doi = {https://doi.org/10.1016/j.knosys.2024.111959},
url = {https://www.sciencedirect.com/science/article/pii/S0950705124005938},
author = {Francesco {Di Fiore} and Laura Mainini},
}

@article{Deepthi2022,
title = {Verification of integrity of deployed deep learning models using {Bayesian} Optimization},
journal = {Knowledge-Based Systems},
volume = {241},
pages = {108238},
year = {2022},
issn = {0950-7051},
doi = {https://doi.org/10.1016/j.knosys.2022.108238},
url = {https://www.sciencedirect.com/science/article/pii/S0950705122000703},
author = {Deepthi Praveenlal Kuttichira and Sunil Gupta and Dang Nguyen and Santu Rana and Svetha Venkatesh},
}

@article{balandat2020botorch,
  title={{BoTorch}: A framework for efficient {Monte-Carlo} {{Bayesian}} optimization},
  author={Balandat, Maximilian and Karrer, Brian and Jiang, Daniel and Daulton, Samuel and Letham, Ben and Wilson, Andrew G and Bakshy, Eytan},
  journal={Advances in neural information processing systems},
  volume={33},
  pages={21524--21538},
  year={2020}
}

@article{baillie2025,
  title={Efficient multi-fidelity {G}aussian process regression for noisy outputs and non-nested experimental designs},
  author={Baillie, Nils and Kerleguer, Baptiste and Feau, Cyril and Garnier, Josselin},
  journal={arXiv preprint arXiv:2511.20183},
  year={2025}
}

@inproceedings{Nogueira2016,
  author       = {João Nogueira and Ruben Martinez-Cantin and Alexandre Bernardino and Leonardo Jamone},
  title        = {Unscented {Bayesian} Optimization for Safe Robot Grasping},
  booktitle    = {2016 IEEE/RSJ International Conference on Intelligent Robots and Systems (IROS)},
  pages        = {1967--1972},
  year         = {2016},
  month        = {October},
  publisher    = {IEEE},
  doi          = {10.1109/IROS.2016.7759282},
  url          = {https://doi.org/10.1109/IROS.2016.7759282}
}

@article{Rai2019,
  author       = {Aravind Rai and Rika Antonova and Franziska Meier and Christopher G. Atkeson},
  title        = {Using Simulation to Improve Sample-Efficiency of {Bayesian} Optimization for Bipedal Robots},
  journal      = {Journal of Machine Learning Research},
  volume       = {20},
  number       = {49},
  pages        = {1--24},
  year         = {2019},
  url          = {http://jmlr.org/papers/v20/18-232.html},
  issn         = {1533-7928}
}

@inproceedings{snoek2012,
 author = {Snoek, Jasper and Larochelle, Hugo and Adams, Ryan P},
 booktitle = {Advances in Neural Information Processing Systems},
 editor = {F. Pereira and C.J. Burges and L. Bottou and K.Q. Weinberger},
 pages = {},
 publisher = {Curran Associates, Inc.},
 title = {Practical {Bayesian} Optimization of Machine Learning Algorithms},
 url = {https://proceedings.neurips.cc/paper_files/paper/2012/file/05311655a15b75fab86956663e1819cd-Paper.pdf},
 volume = {25},
 year = {2012}
}

@inproceedings{Lam2018,
  author       = {R. Lam and M. Poloczek and P. Frazier and K. E. Willcox},
  title        = {Advances in {Bayesian} Optimization with Applications in Aerospace Engineering},
  booktitle    = {2018 AIAA Non-Deterministic Approaches Conference},
  year         = {2018},
  pages        = {1656},
  publisher    = {American Institute of Aeronautics and Astronautics},
  doi          = {10.2514/6.2018-1656},
  url          = {https://doi.org/10.2514/6.2018-1656}
}

@InProceedings{rashidi2024,
  title = 	 {Cylindrical {T}hompson Sampling for High-Dimensional {B}ayesian Optimization},
  author =       {Rashidi, Bahador and Johnstonbaugh, Kerrick and Gao, Chao},
  booktitle = 	 {Proceedings of The 27th International Conference on Artificial Intelligence and Statistics},
  pages = 	 {3502--3510},
  year = 	 {2024},
  volume = 	 {238},
  series = 	 {Proceedings of Machine Learning Research},
  publisher =    {PMLR},
  url = 	 {https://proceedings.mlr.press/v238/rashidi24a.html}
}

@article{palazzolo2026,
      title={Optimal Control of Microswimmers for Trajectory Tracking Using {Bayesian} Optimization}, 
      author={Lucas Palazzolo and Mickaël Binois and Laëtitia Giraldi},
      year={2026},
      eprint={2602.09563},
      journal={arXiv preprint arXiv:2602.09563},
      archivePrefix={arXiv},
      primaryClass={cs.RO},
      url={https://arxiv.org/abs/2602.09563}, 
}

@article{andres2025,
  title={Solar: a solar thermal power plant simulator for blackbox optimization benchmarking},
  author={Andr{\'e}s-Thi{\'o}, Nicolau and Audet, Charles and Diago, Miguel and Gheribi, Aimen E and Digabel, S{\'e}bastien Le and Lebeuf, Xavier and Garneau, Mathieu Lemyre and Tribes, Christophe},
  journal={Optimization and Engineering},
  volume={26},
  number={3},
  pages={1815--1861},
  year={2025},
  publisher={Springer}
}

@article{hernandez2016,
  title={A general framework for constrained {Bayesian} optimization using information-based search},
  author={Hern{\'a}ndez-Lobato, Jos{\'e} Miguel and Gelbart, Michael A and Adams, Ryan P and Hoffman, Matthew W and Ghahramani, Zoubin},
  journal={Journal of Machine Learning Research},
  volume={17},
  number={160},
  pages={1--53},
  year={2016}
}

@article{shinkyu2005,
author = {Jeong, Shinkyu and Murayama, Mitsuhiro and Yamamoto, Kazuomi},
title = {Efficient Optimization Design Method Using Kriging Model},
journal = {Journal of Aircraft},
volume = {42},
number = {2},
pages = {413-420},
year = {2005},
doi = {10.2514/1.6386},
URL = { https://doi.org/10.2514/1.6386
},
eprint = { https://doi.org/10.2514/1.6386
}
}

@InProceedings{mark1989,
author="Drela, Mark",
editor="Mueller, Thomas J.",
title="{XFOIL}: An Analysis and Design System for Low {R}eynolds Number Airfoils",
booktitle="Low {R}eynolds Number Aerodynamics",
year="1989",
publisher="Springer Berlin Heidelberg",
address="Berlin, Heidelberg",
pages="1--12",
isbn="978-3-642-84010-4"
}

@article{dellavecchia2014,
title = {An airfoil shape optimization technique coupling {PARSEC} parameterization and evolutionary algorithm},
journal = {Aerospace Science and Technology},
volume = {32},
number = {1},
pages = {103-110},
year = {2014},
issn = {1270-9638},
doi = {https://doi.org/10.1016/j.ast.2013.11.006},
url = {https://www.sciencedirect.com/science/article/pii/S1270963813002046},
author = {P. Della Vecchia and E? Daniele and E. D'Amato},
}

@Inbook{sobieczky1999,
author="Sobieczky, Helmut",
title="Parametric Airfoils and Wings",
bookTitle="Recent Development of Aerodynamic Design Methodologies: Inverse Design and Optimization",
year="1999",
publisher="Vieweg+Teubner Verlag",
address="Wiesbaden",
pages="71--87",
isbn="978-3-322-89952-1",
doi="10.1007/978-3-322-89952-1_4",
url="https://doi.org/10.1007/978-3-322-89952-1_4"
}

@article{mainini2025,
  author    = {Laura Mainini and Andrea Serani and Hayriye Pehlivan-Solak and 
               Francesco Di Fiore and Markus P. Rumpfkeil and Edmondo Minisci and 
               Domenico Quagliarella and Sihmehmet Yildiz and Simone Ficini and 
               Riccardo Pellegrini and Andrew Thelen and Dean Bryson and 
               Melike Nikbay and Matteo Diez and Philip S. Beran},
  title     = {Analytical benchmark problems and methodological framework for the assessment and comparison of multifidelity optimization methods},
  journal   = {Archives of Computational Methods in Engineering},
  year      = {2025},
  doi       = {10.1007/s11831-025-10392-8},
  url       = {https://doi.org/10.1007/s11831-025-10392-8},
  issn      = {1886-1784}
}

@article{agrawal2025,
doi = {10.1088/2632-2153/ad8e2b},
url = {https://doi.org/10.1088/2632-2153/ad8e2b},
year = {2025},
month = {jan},
publisher = {IOP Publishing},
volume = {6},
number = {1},
pages = {015024},
author = {Agrawal, Atul and Ravi, Kislaya and Koutsourelakis, Phaedon-Stelios and Bungartz, Hans-Joachim},
title = {Stochastic black-box optimization using multi-fidelity score function estimator},
journal = {Machine Learning: Science and Technology}
}

@article{letham2019,
author = {Benjamin Letham and Brian Karrer and Guilherme Ottoni and Eytan Bakshy},
title = {{Constrained {Bayesian} Optimization with Noisy Experiments}},
volume = {14},
journal = {{Bayesian} Analysis},
number = {2},
publisher = {International Society for {Bayesian} Analysis},
pages = {495 -- 519},
year = {2019},
doi = {10.1214/18-BA1110},
URL = {https://doi.org/10.1214/18-BA1110}
}

@article{kandasamy2019,
  title={Multi-fidelity {G}aussian process bandit optimisation},
  author={Kandasamy, Kirthevasan and Dasarathy, Gautam and Oliva, Junier and Schneider, Jeff and Poczos, Barnabas},
  journal={Journal of Artificial Intelligence Research},
  volume={66},
  pages={151--196},
  year={2019}
}

@InProceedings{takeno2020,
  title = 	 {Multi-fidelity {B}ayesian Optimization with Max-value Entropy Search and its Parallelization},
  author =       {Takeno, Shion and Fukuoka, Hitoshi and Tsukada, Yuhki and Koyama, Toshiyuki and Shiga, Motoki and Takeuchi, Ichiro and Karasuyama, Masayuki},
  booktitle = 	 {Proceedings of the 37th International Conference on Machine Learning},
  pages = 	 {9334--9345},
  year = 	 {2020},
  editor = 	 {III, Hal Daumé and Singh, Aarti},
  volume = 	 {119},
  series = 	 {Proceedings of Machine Learning Research},
  month = 	 {13--18 Jul},
  publisher =    {PMLR},
  url = 	 {https://proceedings.mlr.press/v119/takeno20a.html}
}

@article{do2025,
author = {Do, Bach and Zhang, Ruda},
title = {Multifidelity {Bayesian} Optimization: A Review},
journal = {AIAA Journal},
volume = {63},
number = {6},
pages = {2286-2322},
year = {2025},
doi = {10.2514/1.J063812},
URL = {https://doi.org/10.2514/1.J063812},
}

@article{tran2020,
    author = {Tran, Anh and Wildey, Tim and McCann, Scott},
    title = {{sMF-BO-2CoGP}: A Sequential Multi-Fidelity Constrained {Bayesian} Optimization Framework for Design Applications},
    journal = {Journal of Computing and Information Science in Engineering},
    volume = {20},
    number = {3},
    pages = {031007},
    year = {2020},
    month = {04},
    issn = {1530-9827},
    doi = {10.1115/1.4046697},
    url = {https://doi.org/10.1115/1.4046697},
    eprint = {https://asmedigitalcollection.asme.org/computingengineering/article-pdf/20/3/031007/6649818/jcise_20_3_031007.pdf},
}

@article{pang2017,
title = {Discovering variable fractional orders of advection–dispersion equations from field data using multi-fidelity {Bayesian} optimization},
journal = {Journal of Computational Physics},
volume = {348},
pages = {694-714},
year = {2017},
issn = {0021-9991},
doi = {https://doi.org/10.1016/j.jcp.2017.07.052},
url = {https://www.sciencedirect.com/science/article/pii/S0021999117305600},
author = {Guofei Pang and Paris Perdikaris and Wei Cai and George Em Karniadakis},
}

@article{alexandrov1998,
  author  = {Nikolai M. Alexandrov and J. E. Dennis and R. M. Lewis and V. Torczon},
  title   = {A trust-region framework for managing the use of approximation models in optimization},
  journal = {Structural Optimization},
  volume  = {15},
  number  = {1},
  pages   = {16--23},
  year    = {1998},
  doi     = {10.1007/BF01197433},
  issn    = {1615-1488},
  url     = {https://doi.org/10.1007/BF01197433}
}

@article{roustant2020,
author = {Roustant, Olivier and Padonou, Esp\'{e}ran and Deville, Yves and Cl\'{e}ment, Alo\"{\i}s and Perrin, Guillaume and Giorla, Jean and Wynn, Henry},
title = {Group Kernels for {G}aussian Process Metamodels with Categorical Inputs},
journal = {SIAM/ASA Journal on Uncertainty Quantification},
volume = {8},
number = {2},
pages = {775-806},
year = {2020},
doi = {10.1137/18M1209386},
URL = {        https://doi.org/10.1137/18M1209386
},
eprint = {      https://doi.org/10.1137/18M1209386    
}
}

@InProceedings{kandasamy2017,
  title = 	 {Multi-fidelity {B}ayesian Optimisation with Continuous Approximations},
  author =       {Kirthevasan Kandasamy and Gautam Dasarathy and Jeff Schneider and Barnab{\'a}s P{\'o}czos},
  booktitle = 	 {Proceedings of the 34th International Conference on Machine Learning},
  pages = 	 {1799--1808},
  year = 	 {2017},
  editor = 	 {Precup, Doina and Teh, Yee Whye},
  volume = 	 {70},
  series = 	 {Proceedings of Machine Learning Research},
  month = 	 {06--11 Aug},
  publisher =    {PMLR},
  url = 	 {https://proceedings.mlr.press/v70/kandasamy17a.html}
}

@article{cutajar2019,
  title={Deep {G}aussian Processes for Multi-fidelity Modeling},
  author={Kurt Cutajar and Mark Pullin and Andreas C. Damianou and Neil D. Lawrence and Javier I. Gonz{\'a}lez},
  journal={ArXiv},
  year={2019},
  volume={abs/1903.07320},
  url={https://api.semanticscholar.org/CorpusID:81978081}
}

@article{perdikaris2017,
  title={Nonlinear information fusion algorithms for data-efficient multi-fidelity modelling},
  author={Paris Perdikaris and Maziar Raissi and Andreas C. Damianou and Neil D. Lawrence and George Em Karniadakis},
  journal={Proceedings of the Royal Society A: Mathematical, Physical and Engineering Sciences},
  year={2017},
  volume={473},
  number={2198},
  publisher={The Royal Society},
  pages = {20160751},
  url={https://api.semanticscholar.org/CorpusID:4482948}
}

@InProceedings{damianou2013,
  title = 	 {Deep {G}aussian Processes},
  author = 	 {Damianou, Andreas and Lawrence, Neil D.},
  booktitle = 	 {Proceedings of the Sixteenth International Conference on Artificial Intelligence and Statistics},
  pages = 	 {207--215},
  year = 	 {2013},
  editor = 	 {Carvalho, Carlos M. and Ravikumar, Pradeep},
  volume = 	 {31},
  series = 	 {Proceedings of Machine Learning Research},
  address = 	 {Scottsdale, Arizona, USA},
  month = 	 {29 Apr--01 May},
  publisher =    {PMLR},
  url = 	 {https://proceedings.mlr.press/v31/damianou13a.html}
}

@article{legratiet2014,
  author  = {Le Gratiet, Loic and Garnier, Josselin},
  title   = {Recursive co-kriging model for design of computer experiments with multiple levels of fidelity},
  journal = {International Journal for Uncertainty Quantification},
  volume  = {4},
  number  = {5},
  year    = {2014},
  pages   = {},
  url     = {},
  note    = {No page numbers provided}
}

@article{han2012,
author = {Han, Zhong-Hua and G\"{o}rtz, Stefan},
title = {Hierarchical Kriging Model for Variable-Fidelity Surrogate Modeling},
journal = {AIAA Journal},
volume = {50},
number = {9},
pages = {1885-1896},
year = {2012},
doi = {10.2514/1.J051354},
URL = { 
     https://doi.org/10.2514/1.J051354
},
eprint = { 
      https://doi.org/10.2514/1.J051354
}
}

@article{kennedy2000,
  author  = {Michael C. Kennedy and Anthony O'Hagan},
  title   = {Predicting the Output from a Complex Computer Code When Fast Approximations Are Available},
  journal = {Biometrika},
  volume  = {87},
  number  = {1},
  pages   = {1--13},
  year    = {2000},
  url     = {http://www.jstor.org/stable/2673557},
  note    = {Accessed via JSTOR}
}

@InProceedings{eriksson_scalable_2021,
	title = {Scalable Constrained {{Bayesian}} Optimization},
  year         = {2021},
  organization = {PMLR},
  pages        = {730--738},
  author       = {Eriksson, David and Poloczek, Matthias},
    booktitle    = {International Conference on Artificial Intelligence and Statistics}
  
}

@article{papenmeier2025,
  author       = {Leonard Papenmeier and Matthias Poloczek and Luigi Nardi},
  title        = {Understanding High-Dimensional {Bayesian} Optimization},
  year         = {2025},
  archivePrefix = {arXiv},
  eprint       = {2502.09198},
  primaryClass = {stat.ML},
  journal         = {arXiv preprint 2502.09198},
  url          = {https://arxiv.org/abs/2502.09198}
}

@incollection{gramacy2011,
    author = {Gramacy, Robert B. and Lee, Herbert K. H.},
    isbn = {9780199694587},
    title = {Optimization Under Unknown Constraints},
    booktitle = {{Bayesian} Statistics 9},
    publisher = {Oxford University Press},
    year = {2011},
    month = {10},
    doi = {10.1093/acprof:oso/9780199694587.003.0008},
    url = {https://doi.org/10.1093/acprof:oso/9780199694587.003.0008},
    eprint = {https://academic.oup.com/book/0/chapter/141639783/chapter-ag-pdf/45787759/book_1879_section_141639783.ag.pdf},
}

@Book{Rasmussen2006,
  author               = {Rasmussen, Carl E. and Williams, Christopher},
  publisher            = {MIT Press},
  title                = {{Gaussian Processes for Machine Learning}},
  year                 = {2006},
  booktitle            = {Gaussian Processes for Machine Learning},
  url                  = {http://www.gaussianprocess.org/gpml/},
}

@Book{Garnett2023,
  author    = {Garnett, Roman},
  publisher = {Cambridge University Press},
  title     = {{{Bayesian} Optimization}},
  year      = {2023},
}

@book{pourmohamad2021bayesian,
  title={{Bayesian} optimization with application to computer experiments},
  author={Pourmohamad, Tony and Lee, Herbert KH},
  year={2021},
  publisher={Springer}
}

@Book{Gramacy2020,
  author    = {Gramacy, Robert B},
  publisher = {CRC Press},
  title     = {Surrogates: {G}aussian Process Modeling, Design, and Optimization for the Applied Sciences},
  year      = {2020},
}

@Article{Brevault2020,
  author    = {Brevault, Lo{\"\i}c and Balesdent, Mathieu and Hebbal, Ali},
  journal   = {Aerospace Science and Technology},
  title     = {Overview of {G}aussian process based multi-fidelity techniques with variable relationship between fidelities, application to aerospace systems},
  year      = {2020},
  pages     = {106339},
  volume    = {107},
  publisher = {Elsevier},
}

@Article{Sacher2021,
  author    = {Sacher, Matthieu and Le Maitre, Olivier and Duvigneau, R{\'e}gis and Hauville, Fr{\'e}d{\'e}ric and Durand, Mathieu and Lothod{\'e}, Corentin},
  journal   = {International Journal for Uncertainty Quantification},
  title     = {A non-nested infilling strategy for multifidelity based efficient global optimization},
  year      = {2021},
  number    = {1},
  volume    = {11},
  publisher = {Begel House Inc.},
}

@Article{Eriksson2019,
  author  = {Eriksson, David and Pearce, Michael and Gardner, Jacob and Turner, Ryan D and Poloczek, Matthias},
  journal = {Advances in Neural Information Processing Systems},
  title   = {Scalable global optimization via local {Bayesian} optimization},
  year    = {2019},
  pages   = {5496--5507},
  volume  = {32},
}

@Article{Diouane2021,
  author    = {Diouane, Youssef and Picheny, Victor and Riche, Rodolophe Le and Perrotolo, Alexandre Scotto Di},
  journal   = {Journal of Global Optimization},
  title     = {{TREGO}: a trust-region framework for efficient global optimization},
  year      = {2023},
  number    = {1},
  pages     = {1--23},
  volume    = {86},
  publisher = {Springer},
}

@Article{Santoni2024,
  author    = {Santoni, Maria Laura and Raponi, Elena and Leone, Renato De and Doerr, Carola},
  journal   = {ACM Transactions on Evolutionary Learning},
  title     = {Comparison of high-dimensional {Bayesian} optimization algorithms on {BBOB}},
  year      = {2024},
  number    = {3},
  pages     = {1--33},
  volume    = {4},
  publisher = {ACM New York, NY},
}

@inproceedings{Poloczek2017,
 author = {Poloczek, Matthias and Wang, Jialei and Frazier, Peter},
 booktitle = {Advances in Neural Information Processing Systems},
 editor = {I. Guyon and U. Von Luxburg and S. Bengio and H. Wallach and R. Fergus and S. Vishwanathan and R. Garnett},
 pages = {},
 publisher = {Curran Associates, Inc.},
 title = {Multi-Information Source Optimization},
 url = {https://proceedings.neurips.cc/paper_files/paper/2017/file/df1f1d20ee86704251795841e6a9405a-Paper.pdf},
 volume = {30},
 year = {2017}
}

@Article{Gu2018,
  author    = {Gu, Mengyang and Wang, Xiaojing and Berger, James O},
  journal   = {The Annals of Statistics},
  title     = {Robust {G}aussian stochastic process emulation},
  year      = {2018},
  number    = {6A},
  pages     = {3038--3066},
  volume    = {46},
  publisher = {JSTOR},
}

@Article{Marrel2024a,
  author    = {Marrel, Amandine and Iooss, Bertrand},
  journal   = {Reliability Engineering \& System Safety},
  title     = {Probabilistic surrogate modeling by {G}aussian process: A new estimation algorithm for more robust prediction},
  year      = {2024},
  pages     = {110120},
  volume    = {247},
  publisher = {Elsevier},
}

@InProceedings{Mockus1975,
  Title                    = {On {Bayesian} methods for seeking the extremum},
  Author                   = {Mo{\v{c}}kus, J},
  Booktitle                = {Optimization Techniques IFIP Technical Conference},
  Year                     = {1975},
  Organization             = {Springer},
  Pages                    = {400--404},
}

@InProceedings{Srinivas2009,
  author       = {Srinivas, Niranjan and Krause, Andreas and Kakade, Sham and Seeger, Matthias},
  booktitle    = {Proceedings of the 27th International Conference on Machine Learning},
  title        = {{G}aussian Process Optimization in the Bandit Setting: No Regret and Experimental Design},
  year         = {2010},
  organization = {Omnipress},
  pages        = {1015--1022},
}

@Article{Villemonteix2009,
  Title                    = {An informational approach to the global optimization of expensive-to-evaluate functions},
  Author                   = {Villemonteix, Julien and Vazquez, Emmanuel and Walter, Eric},
  Journal                  = {Journal of Global Optimization},
  Year                     = {2009},
  Number                   = {4},
  Pages                    = {509--534},
  Volume                   = {44},
  Publisher                = {Springer},
  Url                      = {http://www.springerlink.com/index/T670U067V47922VK.pdf}
}

@InCollection{Ginsbourger2010a,
  Title                    = {Kriging is well-suited to parallelize optimization},
  Author                   = {Ginsbourger, David and Le Riche, Rodolphe and Carraro, Laurent},
  Booktitle                = {Computational Intelligence in Expensive Optimization Problems},
  Publisher                = {Springer},
  Year                     = {2010},
  Pages                    = {131--162}
}

@article{Thompson1933,
  title = {On the Likelihood that One Unknown Probability Exceeds Another in View of the Evidence of Two Samples},
  volume = {25},
  ISSN = {0006-3444},
  url = {http://dx.doi.org/10.2307/2332286},
  DOI = {10.2307/2332286},
  number = {3/4},
  journal = {Biometrika},
  publisher = {JSTOR},
  author = {Thompson,  William R.},
  year = {1933},
  month = Dec,
  pages = {285}
}

@Article{Binois2022,
  author    = {Binois, Mickael and Wycoff, Nathan},
  journal   = {ACM Transactions on Evolutionary Learning and Optimization},
  title     = {A survey on high-dimensional {G}aussian process modeling with application to {B}ayesian optimization},
  year      = {2022},
  number    = {2},
  pages     = {1--26},
  volume    = {2},
  publisher = {ACM New York, NY},
}

@Article{Dubrule1983,
  Title                    = {Cross validation of kriging in a unique neighborhood},
  Author                   = {Dubrule, Olivier},
  Journal                  = {Mathematical Geology},
  Year                     = {1983},
  Number                   = {6},
  Pages                    = {687--699},
  Volume                   = {15},
  Publisher                = {Springer}
}

@Article{LeDigabel2023,
  author    = {Le Digabel, S{\'e}bastien and Wild, Stefan M},
  journal   = {Optimization and Engineering},
  title     = {A taxonomy of constraints in black-box simulation-based optimization},
  year      = {2023},
  pages     = {1--19},
  publisher = {Springer},
}

@InProceedings{Swersky2013,
  author    = {Swersky, Kevin and Snoek, Jasper and Adams, Ryan P},
  title     = {Multi-task {Bayesian} optimization},
  booktitle = {Advances in neural information processing systems},
  year      = {2013},
  pages     = {2004--2012},
}

@inproceedings{tighineanu2022transfer,
  title={Transfer learning with {G}aussian processes for {Bayesian} optimization},
  author={Tighineanu, Petru and Skubch, Kathrin and Baireuther, Paul and Reiss, Attila and Berkenkamp, Felix and Vinogradska, Julia},
  booktitle={International conference on artificial intelligence and statistics},
  pages={6152--6181},
  year={2022},
  organization={PMLR}
}

\end{document}